%% file: pdgs_full.tex
\documentclass[lettersize,journal]{IEEEtran}
\usepackage{amsmath,amsfonts}
\usepackage{algorithmic}
\usepackage{algorithm}
\usepackage{array}
\usepackage[caption=false,font=normalsize,labelfont=sf,textfont=sf]{subfig}
\usepackage{textcomp}
\usepackage{stfloats}
\usepackage{url}
\usepackage{verbatim}
\usepackage{graphicx}
\usepackage{cite}
\usepackage{multirow}
\usepackage{colortbl}
\usepackage{xcolor}
\usepackage{etoolbox}
\usepackage{caption}
\definecolor{c1}{HTML}{ffcc99}
\definecolor{c2}{HTML}{fff8ae}
\newcommand{\ye}[1]{\textcolor[rgb]{0.00,0.0,0.0}{#1}}

\newcommand{\figref}[1]{Fig.~\ref{#1}}
\newcommand{\tabref}[1]{Table~\ref{#1}}
\newcommand{\equref}[1]{Eq.~(\ref{#1})}

\newcommand{\secref}[1]{Sec.~\ref{#1}}

\begin{document}

\title{Progressive Radiance Distillation for Inverse Rendering with Gaussian Splatting}

\author{Keyang Ye, Qiming Hou, and Kun Zhou
\IEEEcompsocitemizethanks{\IEEEcompsocthanksitem 
K. Zhou is the corresponding author. All authors are with State Key Lab of CAD \& CG, Zhejiang University, Hangzhou 310058, China.\\
E-mail: kunzhou@acm.org
}}

\markboth{Journal of \LaTeX\ Class Files,~Vol.~14, No.~8, August~2021}%
{Shell \MakeLowercase{\textit{et al.}}: A Sample Article Using IEEEtran.cls for IEEE Journals}


\newcommand{\insertfig}{\includegraphics[width=\linewidth]{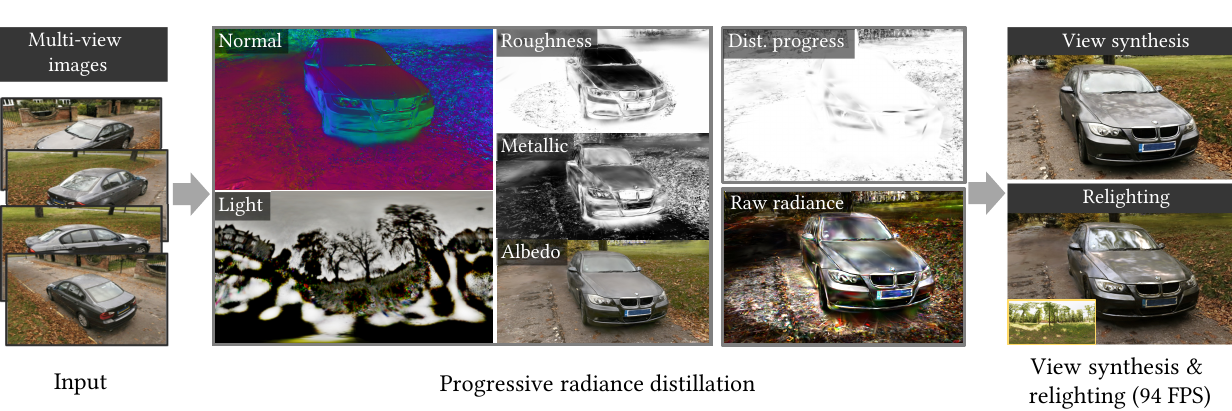}\captionof*{figure}{We propose progressive radiance distillation for geometry-light-material decomposition from multi-view images. Our rendering model combines physically-based rendering with Gaussian-based radiance field rendering using a distillation progress map, which outperforms state-of-the-art methods in both novel view synthesis and relighting. The scene is \textit{sedan} from~\cite{ref_nerf}, a real-world capture. We use a spherical mask to restrict distillation to foreground objects, which is visible in the distilled layers but does not manifest as noticeable artifacts in final results.}}

\makeatletter
\apptocmd{\@maketitle}{\centering\insertfig}{}{}
\makeatother

\maketitle

\begin{abstract}

We propose \emph{progressive radiance distillation}, an inverse rendering method that combines physically-based rendering with Gaussian-based radiance field rendering using a distillation progress map. Taking multi-view images as input, our method starts from a pre-trained radiance field guidance, and distills physically-based light and material parameters from the radiance field using an image-fitting process. 
The distillation progress map is initialized to a small value, which favors radiance field rendering. During early iterations when fitted light and material parameters are far from convergence, the radiance field fallback ensures the sanity of image loss gradients and avoids local minima that attracts under-fit states. As fitted parameters converge, the physical model gradually takes over and the distillation progress increases correspondingly. In presence of light paths unmodeled by the physical model, the distillation progress never finishes on affected pixels and the learned radiance field stays in the final rendering. With this designed tolerance for physical model limitations, we prevent unmodeled color components from leaking into light and material parameters, alleviating relighting artifacts. Meanwhile, the remaining radiance field compensates for the limitations of the physical model, guaranteeing high-quality novel views synthesis. Experimental results demonstrate that our method significantly outperforms state-of-the-art techniques quality-wise in both novel view synthesis and relighting. The idea of progressive radiance distillation is not limited to Gaussian splatting. We show that it also has positive effects for prominently specular scenes when adapted to a mesh-based inverse rendering method.

\end{abstract}


\section{Introduction}
\IEEEPARstart{I}{nverse} rendering, a pivotal discipline within computer graphics and vision, solves the rendering equation~\cite{render_equation} to decompose observed images into physically-based components such as lighting conditions, material properties, and geometry. When interpreted as a regression problem, though, the rendering equation poses significant mathematical ambiguity. Light and material can be challenging to factorize as they always appear in the same integration and infinitely many combinations can yield the same pixel colors. Such ambiguity frequently allows approximation errors from idealized light-material assumptions to leak into model parameters, which can cause visible artifacts when synthesizing new images.


This ambiguity problem can be side-stepped by dropping the light-material decomposition in favor of a \emph{radiance field}, a learned function parameterized over spatial position and view direction. Such a function is significantly more robust to learn, leading to high-fidelity novel view synthesis. This has been demonstrated by the seminal works of NeRF (Neural Radiance Field) \cite{nerf} and 3DGS (3D Gaussian Splatting) \cite{3DGS}. The robustness comes at a cost, though, as light and material remain entangled and cannot change independently during image synthesis.

Many attempts \cite{nerfactor,nerd,nero,jiang2023gaussianshader,liang2024gsir} have been made to apply successful radiance field function representations to inverse rendering, among which several approaches \cite{jiang2023gaussianshader, GIR} keep the original uninterpreted radiance alongside light-material integration as an indirect light approximation and train their models from scratch. Such a radiance term alleviates novel view synthesis quality issues caused by the limitations of physically-based rendering models.
However, function representations are orthogonal to the ambiguity problem and additive uninterpreted radiance amplifies it by adding yet another ambiguity source.


We tackle the ambiguity problem by treating radiance fields and physically-based models as interchangeable rather than complimentary. \ye{Specifically, we relate them analogously to the \emph{distillation} concept in physics, where different components are gradually extracted from a liquid as the temperature rises. Compared to joint optimization from scratch, which attempts to factorize all components at once, distillation effectively isolates the ambiguities between different components, resulting in a more accurate decomposition.}

We propose \emph{progressive radiance distillation}, an inverse rendering method using Gaussian splatting. It initializes from a pre-trained Gassian-based radiance field guidance, and distills physically-based light and material parameters from the radiance field using a hybrid image-fitting process. Specifically, for each training image, we render one image using radiance field and another using physically-base parameters. The loss function is computed on a final predicted image, linearly interpolated from the two renderings using a learned \emph{distillation progress map} for weights, which guarantees an image quality better or equal to the raw radiance baseline by optimizing the radiance distribution.

The distillation progress map is initialized to a small value, which favors the radiance field and only introduces a minuscule amount of physically-based rendering. During early iterations when fitted physical parameters are still far from convergence, the radiance field fallback ensures the sanity of image loss gradients and avoids local minima that attracts under-fit states.
As light and material parameters converge to a comparable level of fitting accuracy, the physical model gradually takes over and the distillation progress increases correspondingly. 
In presence of light paths unmodeled by the physical model, the distillation progress never reaches 100\% on affected pixels and the learned radiance field stays in the final rendering. With this designed tolerance for physical model limitations, we prevent unmodeled color components from leaking into light and material parameters, alleviating ambiguity-related artifacts for better relighting. Meanwhile, the remaining radiance field compensates for the limitations of the physical model, guaranteeing high-quality novel views synthesis.
To further reduce ambiguity, we choose to fit specular and diffuse parameters separately as opposed to the more common joint optimization. In presence of the radiance field fallback, this allows each optimization stage to focus on the pixels where the fitted components are most prominent. We choose to fit the specular component first as its physical model out-competes learned radiance faster, making it easier to distill.

Experimental results on several published datasets demonstrate that our method significantly outperforms state-of-the-art techniques quality-wise in both novel view synthesis and relighting, especially enhancing the reconstruction accuracy for scenes with prominent specular reflections. Ablation studies shows that our stage-wise radiance distillation is effective at avoiding early-stage local minima and alleviates light-material ambiguities. We also explore the possibility of applying our progressive distillation paradigm to \ye{mesh-based} methods, demonstrating its generalization potential for inverse rendering.

\begin{figure*}[t]
\centering
\includegraphics[width=0.98\linewidth]{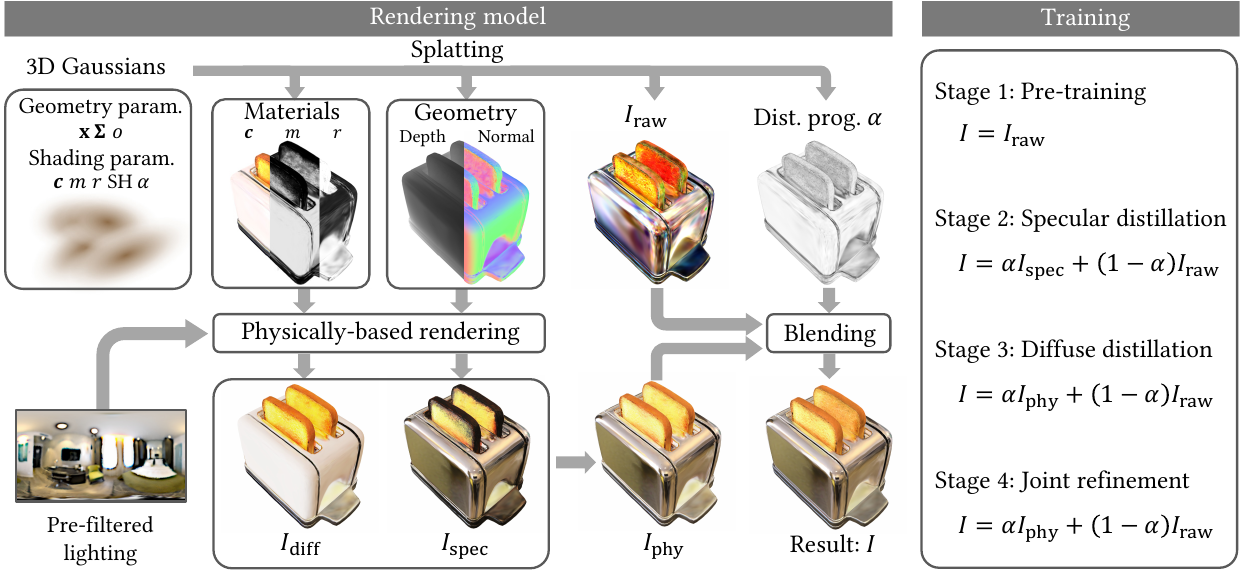}
\caption{The rendering model and training stages of our method. We adopt a deferred shading pipeline. First, geometry parameters (position: $\mathbf{x}$, covariance: $\mathbf{\Sigma}$ and opacity: $o$) and shading parameters (albedo: $\boldsymbol{c}$, metallic: $m$, roughness: $r$, SH colors $y_l^m$ and distillation progress: $\alpha$) are splatted into several screen space maps, including material maps (albedo, metallic and roughness), geometry maps (depth and normal), the raw radiance map $\mathrm{I_{raw}}$ and the distillation progress map $\alpha$. Then, we send the maps into a physically-based shader to produce diffuse and specular color maps: $\mathrm{I_{diff}}$ and $\mathrm{I_{spec}}$. The physical term $\mathrm{I_{phy}}$ combines $\mathrm{I_{diff}}$ and $\mathrm{I_{spec}}$, and then gets blended with $\mathrm{I_{raw}}$ using $\alpha$ as weight to produce the final result. The various shading parameters are distilled from $\mathrm{I_{raw}}$ using a four-stage training process. To mitigate ambiguity, we subset and specialize the rendering model for each individual stage.
 \label{fig:pipeline}}
\end{figure*}

\section{Related work}

\ye{Inverse rendering~\cite{intrinsicImage,RaviInverseRendering,SatoInverseRendering}, the task of decomposing the appearance of objects in observed images into the underlying geometry, material properties, and lighting conditions, has been extensively explored in both computer graphics and vision. Making a full decomposition involves two important considerations: how to represent geometry and appearance, and how to address the ill-posed problem.}

\ye{As a widely used geometry representation in graphics pipelines and game engines, meshes represent explicit surfaces and support high performance rendering, which sparks a surge of research in mesh-based differentiable rendering~\cite{InverseRenderingSurvey03, kato2020differentiable, azinovic2019inverse, ndrmc}. However, due to the depth discontinuities that arise when rendering a mesh to an image, and the difficulty posed in changing their topology and avoiding self-intersections~\cite{nicolet2021large}, mesh-based methods require good geometry initialization, have limited ability for fine-tuning inaccurate topology~\cite{luan2021unified}, and are unstable to the ambiguity caused by surface reflections~\cite{queau2018variational}. To alleviate these issues, implicit representations, typified by signed distance field (SDF), have emerged as alternatives~\cite{DVR, IDR}. SDF is typically parameterized by a Multi-Layer Perceptron (MLP), with its zero level set corresponding to the geometric surface. Due to its continuity and independence from topological constraints, shape optimization becomes more flexible. However, the computation of SDF-based ray-surface intersection usually relies on time-consuming ray marching, leading to a long training time. Additionally, SDF tends to provide smooth normals, failing to capture some subtle details. NDR~\cite{nvidiffrec} and Neural-PBIR~\cite{pbir} use SDF as a geometry initialization, extract coarse meshes from the SDF, and fine-tune the meshes through differentiable rendering. They achieve high reconstruction quality on diffuse-dominant objects with a moderate training time while still struggle with reflective objects. They tend to use fragmented geometry within the surface to fit view-dependent changes. The fixed topology limits further fine-tuning after geometry extraction.}

\ye{Similar to SDF, radiance fields methods, such as NeRF~\cite{nerf} and 3DGS~\cite{3DGS}, also construct a implicit and optimizable geometry. NeRF-based methods~\cite{nerfactor, nerv, nerd, physg, zhang2022modeling, mai2023neural} replace the emissive radiance with material properties predicted by MLPs and build a fully differentiable rendering pipeline. However, the predicted volume density is underconstrained, leading to inaccurate normal reconstruction. Some methods~\cite{neus, 
unisurf, nero} replace the density with signed distance and perform volume rendering near the zero level set, but still remain the same shortcomings as SDF-based methods. Recently, 3D Gaussians became a popular representation of radiance fields. Gaussian-based inverse rendering methods~\cite{jiang2023gaussianshader,liang2024gsir,GIR,gao2023relightable,wu2024deferredgs} inherit the highly efficient training and rendering from vanilla 3DGS, but they are difficult to perfectly reconstruct reflective surfaces. 3DGS-DR~\cite{our2024sig3dgsdf} proposed a deferred rendering pipeline with normal propagation, achieving detailed normal reconstruction for reflective surfaces. We choose 3D Gaussians as our geometry representation and incorporate contributions from 3DGS-DR, achieving accurate and efficient reconstruction for objects with various roughness. }

\ye{For the appearance, we choose the widely used Cook-Torrance microfacet model~\cite{cook1982reflectance} with albedo, rougheness and metallic as its parameters. To keep the real-time performance, our rendering is based on image based lighting~\cite{ibr}. Other rendering methods, such as Monte-Carlo path tracing~\cite{whitted1980improved}, adopted by NDRMC~\cite{ndrmc} and Neural-PBIR~\cite{pbir}, may achieve a more thorough decomposition but at the cost of decreased efficiency.}

\ye{The inherent ambiguity has been a recurrent problem since the beginning of inverse rendering research. Most previous methods attempt to address this difficulty by adding priors and regularizations. Specifically, these methods minimize the loss function, which consists of a image loss and other regularization terms, to optimize a physically based rendering model from scratch. In contrast, we alleviate ambiguities by carefully controlling the training process to optimize different components in isolation. To this end, we design a hybrid model linearly combining radiance fields with physical models using optimizable distillation progress as the final rendering model, and gradually increase the distillation progress to drive a full factorization using image-fitting. Similarly, some prior works also utilize a trained radiance field to reconstruct material/lighting parameters. For example, \cite{pbir} initializes the BRDF parameters by fitting the trained radiance field to accelerate convergence. \cite{nimier2022unbiased} recover density from the trained radiance field to initialize the optimization of physical medium parameters. However, they only use the radiance field to initialize the purely physically based model optimization. Besides, some prior works~\cite{jiang2023gaussianshader, GIR} also combine a emissive radiance term with the physical model, regarding it as an additive indirect lighting approximation. However, they are directly added together and jointly optimized from scratch, which can potentially increase ambiguity. In contrast, our distillation process starts from a trained radiance field model, and our linear interpolation design allows the radiance term to remain stable until its weight decreases to nearly zero. Compared with state-of-the-art methods, our hybrid rendering model and progressive distillation mechanism produce superior results in both novel view synthesis (NVS) and relighting, contributing valuably to inverse rendering.}

\section{Progressive Radiance Distillation}
\label{sec:method}




\subsection{Rendering Model}\label{sec:model}

The general workflow of 3DGS~\cite{3DGS} consists of fitting a set of spatial Gaussians carrying learned parameters. Images are synthesized by splatting and blending Gaussian properties to pixels, and the same image synthesis model is used for training and rendering. We extend the pipeline for inverse rendering while following the general workflow.

We categorize the per-Gaussian parameters into \emph{geometry parameters} and \emph{shading parameters}. Geometry parameters consist of the position, rotation, scaling and opacity of each Gaussian. When combined, they completely define the Gaussian-to-pixel mapping, a linear function that splats and blends the per-Gaussian shading parameters to screen space. We then compute the final pixel colors from the splatted screen space parameters using deferred shading~\cite{DeferredRendering}.


We will focus on the final shading step as it is most relevant to our paper. Our full rendering model, the version applied after inverse rendering converges, is specified:

\newcommand{\x}[0]{x}
\newcommand{\wo}[0]{\boldsymbol{\omega}_o}
\newcommand{\wi}[0]{\boldsymbol{\omega}_i}
\newcommand{\wR}[0]{\boldsymbol{\omega}_r}

\begin{equation}
    \mathrm{I}(\x,\wo)=\alpha_x\mathrm{I_{phy}}(\x,\wo)+(1-\alpha_x)\mathrm{I_{raw}}(\x,\wo)
    \label{equ:fullModel}.
\end{equation}

Here $\mathrm{I}(\x,\wo)$ is the final radiance parameterized over spatial position $\x$ and view direction $\wo$. $\mathrm{I_{phy}}(\x,\wo)$ is the physically-based term and $\mathrm{I_{raw}}(\x,\wo)$ is the radiance field term, which we fine-tune during our distillation. $\alpha_x$ is the distillation progress value. It is a learned function of position $x$ and we put the variable in the subscript for clarity. All similar subscripts follow this meaning.

The raw radiance term follows the original 3DGS paper~\cite{3DGS} and uses an order-4 SH (Spherical Harmonics)~\cite{PRT} function:

\begin{equation}
    \mathrm{I_{raw}}(\x,\wo)=\sum_{l=0}^3 \sum_{m=-l}^l y_{l,x}^m \mathrm{Y}_l^m(\wo),
    \label{equ:raw}
\end{equation}
where $\mathrm{Y}_l^m(\wo)$ is the spherical harmonic basis function. $y_{l,x}^m$ are the corresponding coefficients. Note that the SH series are evaluated on the view direction $\wo$.

The physically-based term is based on the Cook-Torrance microfacet model~\cite{cook1982reflectance}:

\begin{equation}
    \mathrm{I_{phy}}(\x,\wo)=(1-m_x)\mathrm{I_{diff}}(\x,\wo)+\mathrm{I_{spec}}(\x,\wo),
    \label{equ:physical}
\end{equation}

\begin{equation}
    \mathrm{I_{diff}}(\x)=\frac{\boldsymbol{c}_x}{\pi} \int_\Omega \mathrm{V}(\x,\wi)\mathrm{L}(\wi) (\mathbf{n}_x \cdot \wi) d \wi,
    \label{equ:diffuse}
\end{equation}

\begin{equation}
    \mathrm{I_{spec}}(\x,\omega_o)=\int_\Omega \mathrm{\rho}(\wi,\wo; \boldsymbol{c}_x, r_x, m_x) \mathrm{L}(\wi) (\mathbf{n}_x \cdot \wi) d \wi,
    \label{equ:specular}
\end{equation}
where $\boldsymbol{c}_x$, $r_x$, $m_x$ are the diffuse albedo map, roughness map and metallic map, respectively. $\mathrm{L}(\wi)$ is the learned incident radiance from incoming direction $\wi$, which we approximate as position-independent. $\mathbf{n}_x$ is the surface normal at $\x$. $\mathrm{V}(\x,\wi)$ is a baked visibility term and $\mathrm{\rho}(\wi,\wo;\boldsymbol{c},r,m)$ is the microfacet BRDF (Bidirectional Reflectance Distribution Function).

All learned functions presented of position $x$ are implemented as per-Gaussian shading parameters. The light term $\mathrm{L}(\wi)$ is implemented as an environment map. The diffuse-term integration is approximated as a SH triple product~\cite{sloan2008stupid}, during which $\mathrm{L}(\wi)$ is projected to SH. The visibility $\mathrm{V}(\x,\wi)$ is pre-computed over a regular grid, with each cell computing a set of SH coefficients by projecting a splatted opacity cube-map. During training, we compute $\mathrm{V}$ once during initialization and keep it frozen throughout the iterations. While we fine-tune geometric parameters, we focus on shading and the visibility changes during training are minimal. As our visibility approximation is low frequency, we opt to drop it from the specular term. The specular component is computed using a split-sum process~\cite{hill2020physically} with pre-filtered environment light,
detailed in the supplement material.

\subsection{Training}
\label{sec:train}

We split training into four stages: pre-training, specular distillation, diffuse distillation, and joint refinement. Each stage is an independent SGD (Stochastic Gradient Descent) loss-minimization session run for a fixed number of epoches. All learned parameters are shared throughout all stages, though each stage may freeze a subset. All stages share an image loss term $\mathcal{L}_\mathrm{rgb}$ identical to baseline 3DGS defined on final radiance $\mathrm{I}(\x,\wo)$. For simplicity, we omit the base image loss and focus our discussion on additional, stage-specific loss terms.

\paragraph{Stage 1: Pre-training.} In this stage, we fix $\alpha_x=0$ and only $\mathrm{I_{raw}}$ is trained. This stage is identical to the 3DGS baseline. We effectively have:
\begin{equation}
    \mathrm{I}(\x,\wo)=\mathrm{I_{raw}}(\x,\wo).
\end{equation}
At the end of this stage, the basic Gaussian geometry and raw radiance are close to convergence, though we keep fine-tuning geometry parameters such as Gaussian rotation, scale and opacity in the following stages.


\paragraph{Stage 2: Specular distillation.} With the baseline $\mathrm{I_{raw}}$ trained, we bump $\alpha_x$ to 0.01 as initialization and fix $m_x=1$ to start fitting specular effects. We distill the specular term first because it is less ambiguous alongside $\mathrm{I_{raw}}$ and takes over faster. The output radiance in this stage is effectively:
\begin{equation}
    \mathrm{I}(\x,\wo)=\alpha_x\mathrm{I_{spec}}(\x,\wo)+(1-\alpha_x)\mathrm{I_{raw}}(\x,\wo).
    \label{equ:stage2}
\end{equation}
Note that $m_x=1$ completely disables the diffuse term, as indicated by \equref{equ:physical}. We follow the normal propagation and color sabotage steps in 3DGS-DR~\cite{our2024sig3dgsdf} for normal reconstruction. Since the physically-based specular reflection approximates high-frequency view-dependent effects better than the SH-based $\mathrm{I_{raw}}$, the alpha map naturally increases in low-roughness areas with solely an image-fitting loss. At the end of this stage, normal and roughness of specular objects, plus the portion of light environment map visible under specular reflection, are essentially distilled to convergence.

\begin{figure}[t]
\centering
\includegraphics[width=1.0\columnwidth]{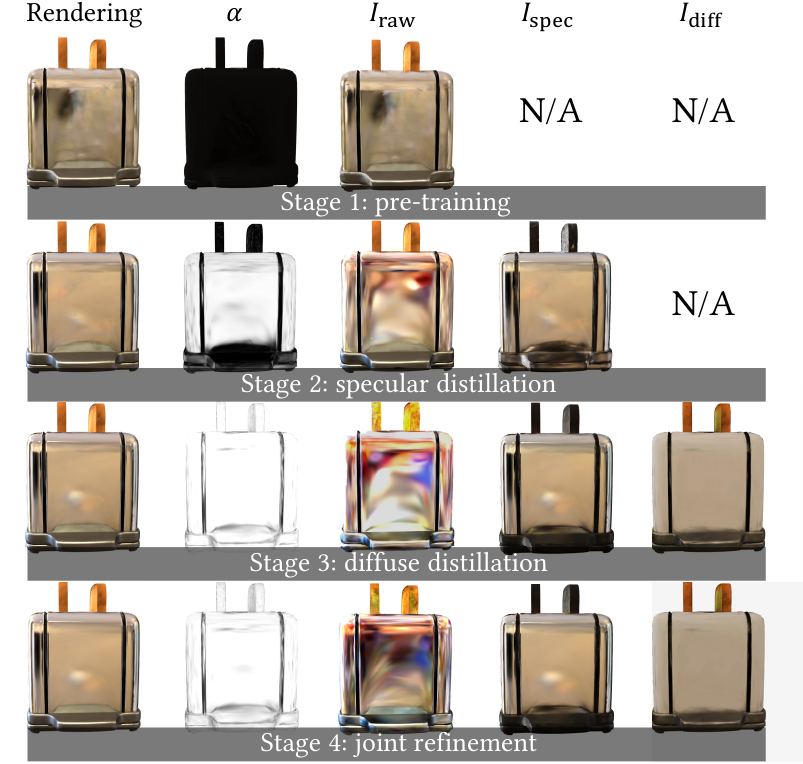}\\
\caption{Visualization of the distillation progress map and different rendering components generated by each optimization stage. \label{fig:stage}}
\end{figure}

The second row of \figref{fig:stage} illustrates the effect of a converged specular distillation. At this point, the diffuse reflection remains in $\mathrm{I_{raw}}$ and the distillation progress on diffuse-only pixels remains close to 0. With the improved specular fitting, the overall model already demonstrates improved novel view synthesis capabilities. However, to further decompose material parameters for relighting, we still need to distill the diffuse term.

Before starting the next stage, we bake the visibility term $\mathrm{V}(\x,\wi)$ using the current Gaussian geometry.

\paragraph{Stage 3: Diffuse distillation.} We unfreeze $m_x$ in this stage and the final output radiance matches the full model in \equref{equ:fullModel}. However, unlike the specular stage, the image loss alone is insufficient to differentiate the diffuse reflection from the similarly low-frequency raw SH radiance. As a solution, we introduce an extra loss term $\mathcal{L}_\alpha$ to promote physical modeling:
\begin{equation}
    \mathcal{L}_\alpha = \mathrm{MSE}(\mathrm{I_{mask}}, \ye{\alpha_x}),
    \label{eq:alpha_loss}
\end{equation}
where $\mathrm{I_{mask}}$ is an object mask. $\mathrm{I_{mask}}$ is 1 on pixels covered by any object and 0 otherwise. Note that our method does not require per-object masks. We describe how to handle real-world data with no readily-available masks in the supplementary material. With a soft constraint pulling $\alpha$ to 1, explainable portions of raw radiance are distilled into diffuse reflection, and specular reflection adapts to the changed $m_x$ accordingly. At the end of this stage, raw radiance naturally shift to unmodeled light paths such as inter-reflection for high-$\alpha$ pixels.

We follow previous works~\cite{nvidiffrec, GIR} and add $\mathcal{L}_\mathrm{light}$ to penalize coloration and make the light environment map as close to neutral white as possible. This is achieved by minimizing the difference between individual RGB channel values and their mean using the following equation:

\begin{equation}
    \mathcal{L}_\mathrm{light} =\sum_{i=1}^3|l_i-\frac{1}{3}\sum_{j=1}^3 l_j|,
\end{equation}
where $l_i$ and $l_j$ represent the individual RGB components of the light. The final loss function for SGD is $\mathcal{L}=\lambda_1\mathcal{L}_\mathrm{rgb}+\lambda_2\mathcal{L}_\alpha+\lambda_3\mathcal{L}_\mathrm{light}$, where \ye{$\lambda_1=1.0, \lambda_2=0.08, \lambda_3=0.003$} are tunable meta-parameters.



\paragraph{Stage 4: Joint refinement.} In this final stage, we slightly reduce the learning rate of $\alpha$ and fine-tune all other parameters. For real scenes, we also freeze the light $L$ and drop the $\mathrm{I_{mask}}$ term from $\mathcal{L}_\alpha$:
\begin{equation}
    \mathcal{L}_\alpha = \mathrm{MSE}(1, \ye{\alpha_x}),
    \label{equ:realScene}
\end{equation}
Note that switching to \equref{equ:realScene} would distill the entire image, including background pixels.

The real scene-specific designs are motivated by relighting quality, as we prefer background pixels to respond consistently with foreground pixels in presence of light changes. To achieve that, we need to encourage a diffuse fitting on the previously neglected background pixels. However, such pixels are poorly explained by our shading model and we have to anticipate high fitting errors. We therefore freeze the light $L$ to cut off the main gradient path that facilitates cross-pixel error propagation, which limits background fitting errors to their own pixels.



\section{Results and Evaluation}

\begin{table*}[t]
\caption{Dataset-averaged image quality comparison for novel views synthesis.\label{tab:novel_view_cp}}
\tabcolsep=0.21cm
\renewcommand\arraystretch{1.2}
\input{tabs/novel_view_cp}
\end{table*}

\begin{figure*}[t]
\centering
\includegraphics[width=\linewidth]{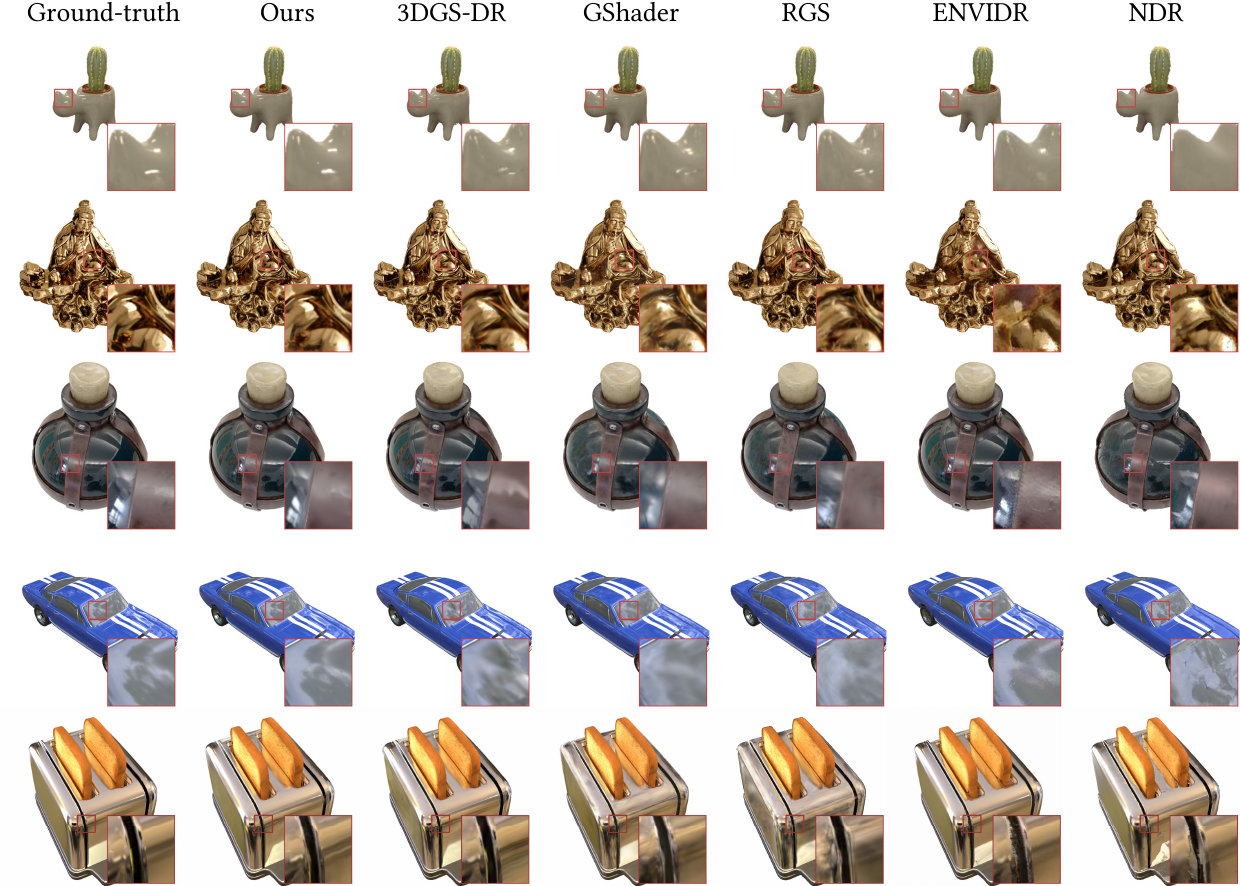}
\caption{Qualitative comparisons of novel view synthesis results. From top to bottom: cactus (from Stanford ORB~\cite{stanfordorb}), luyu and potion (from Glossy Synthetic~\cite{nero}), car and toaster (from Shiny Blender~\cite{ref_nerf}). \label{fig:novel_view}}
\end{figure*}

\begin{figure*}
\centering
\includegraphics[width=\linewidth]{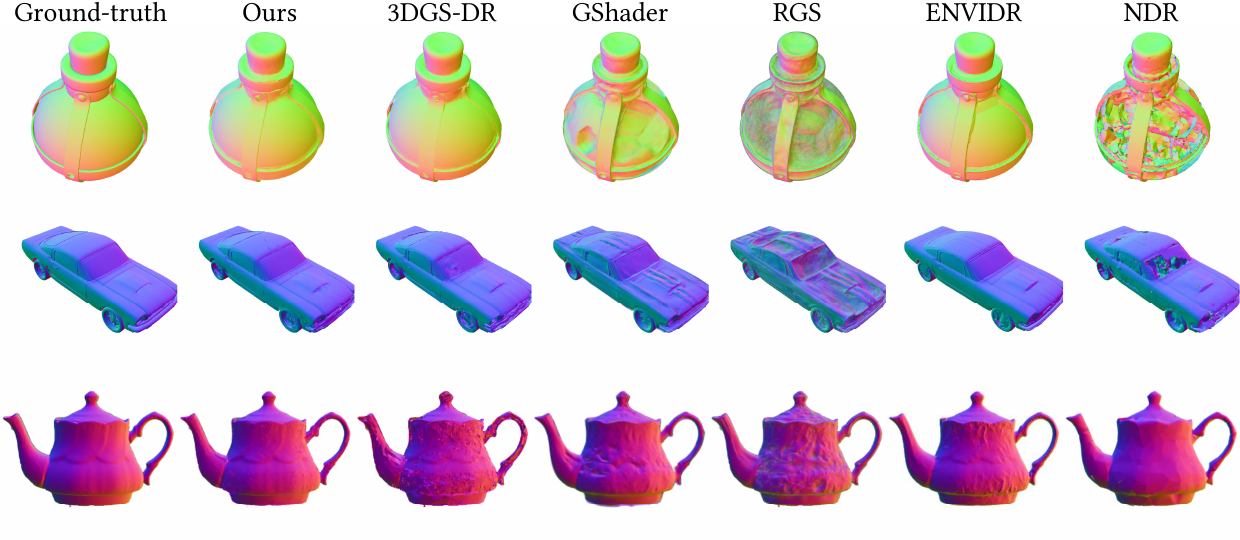}
\caption{Qualitative comparisons of normal estimated by different methods. From top to bottom: potion (from Glossy Synthetic~\cite{nero}), car (from Shiny Blender~\cite{ref_nerf}) and teapot (from Stanford ORB~\cite{stanfordorb}). \label{fig:normal_cp}}
\end{figure*}

\begin{figure*}
\centering
\includegraphics[width=\linewidth]{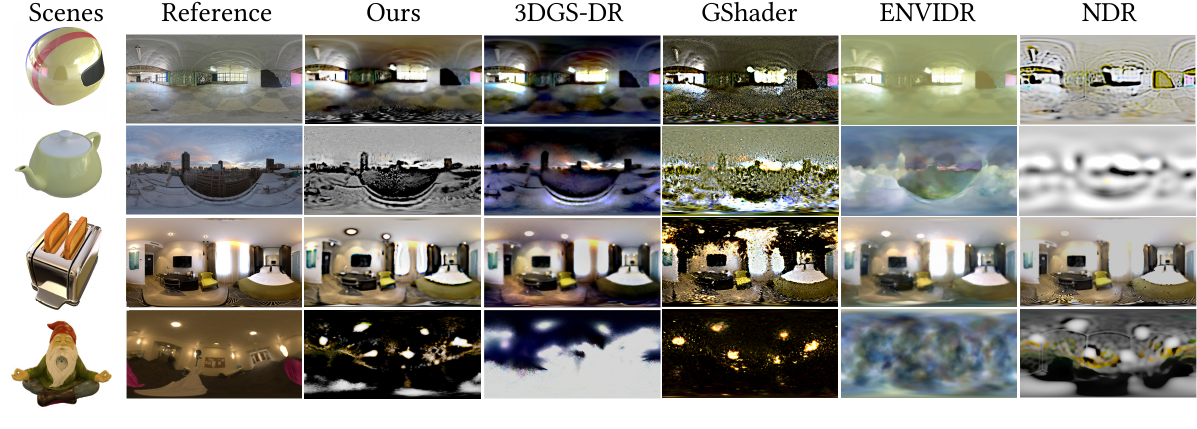}
\caption{ Qualitative comparisons of environment maps estimated by different methods. From top to bottom: helmet, teapot and toaster (from Shiny Blender~\cite{ref_nerf}), gnome (from Stanford ORB~\cite{stanfordorb}).\label{fig:envlight_cp}}
\end{figure*}

\begin{figure*}
\centering
\includegraphics[width=\linewidth]{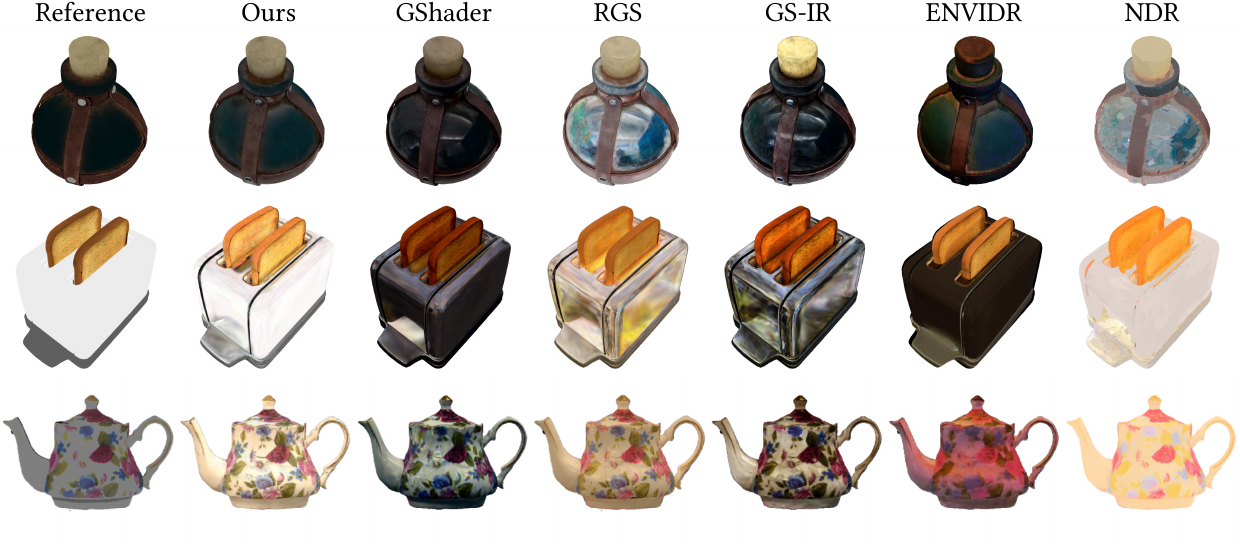}
\caption{Qualitative comparisons of albedo estimated by different methods. Due to differences in rendering models among different methods, as well as the inherent ambiguity of light and material, the comparison is only for reference. From top to bottom:potion (from Glossy Synthetic~\cite{nero}), toaster (from Shiny Blender~\cite{ref_nerf}) and teapot (from Stanford ORB~\cite{stanfordorb}). \label{fig:albedo_cp}}
\end{figure*}

\begin{table}[t]
\caption{ Normal and light reconstruction quality (evaluated
by MAE$^\circ$ and LPIPS respectively) comparisons on the Shiny
Blender Dataset.\label{tab:normal_env_cp}}
\tabcolsep=0.18cm
\renewcommand\arraystretch{1.2}
\input{tabs/normal_env_cp}
\end{table}

\begin{figure*}
\centering
\includegraphics[width=\linewidth]{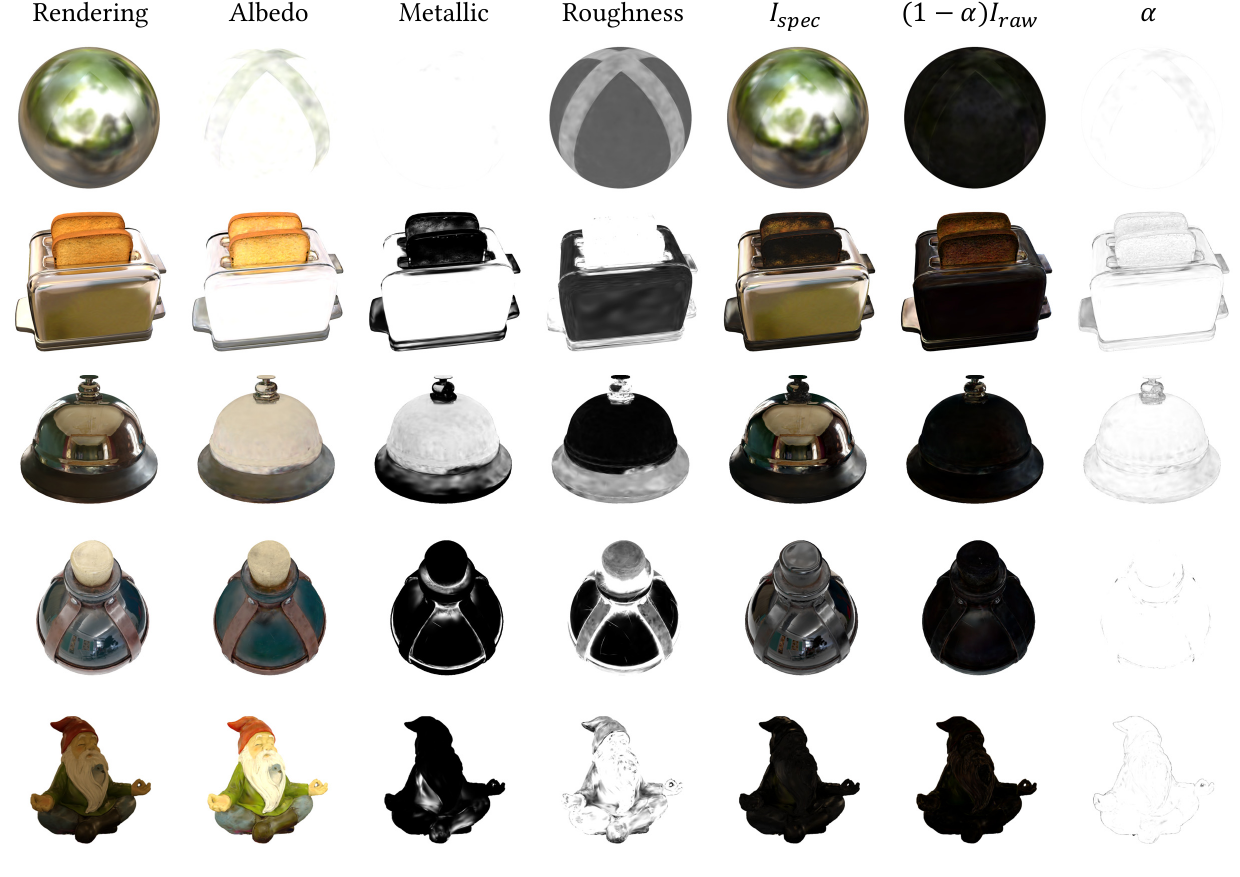}
\caption{Decomposition results of our method. We display $(1-\alpha)\mathrm{I_{raw}}$ instead of $\mathrm{I_{raw}}$ to help illustrate its contribution to final results. From top to bottom: ball and toaster (from Shiny Blender~\cite{ref_nerf}), tbell and potion (from Glossy Synthetic~\cite{nero}), gnome (from Stanford ORB~\cite{stanfordorb}). \label{fig:decomp}}
\end{figure*}

\begin{figure*}[!t]
\centering
\includegraphics[width=\linewidth]{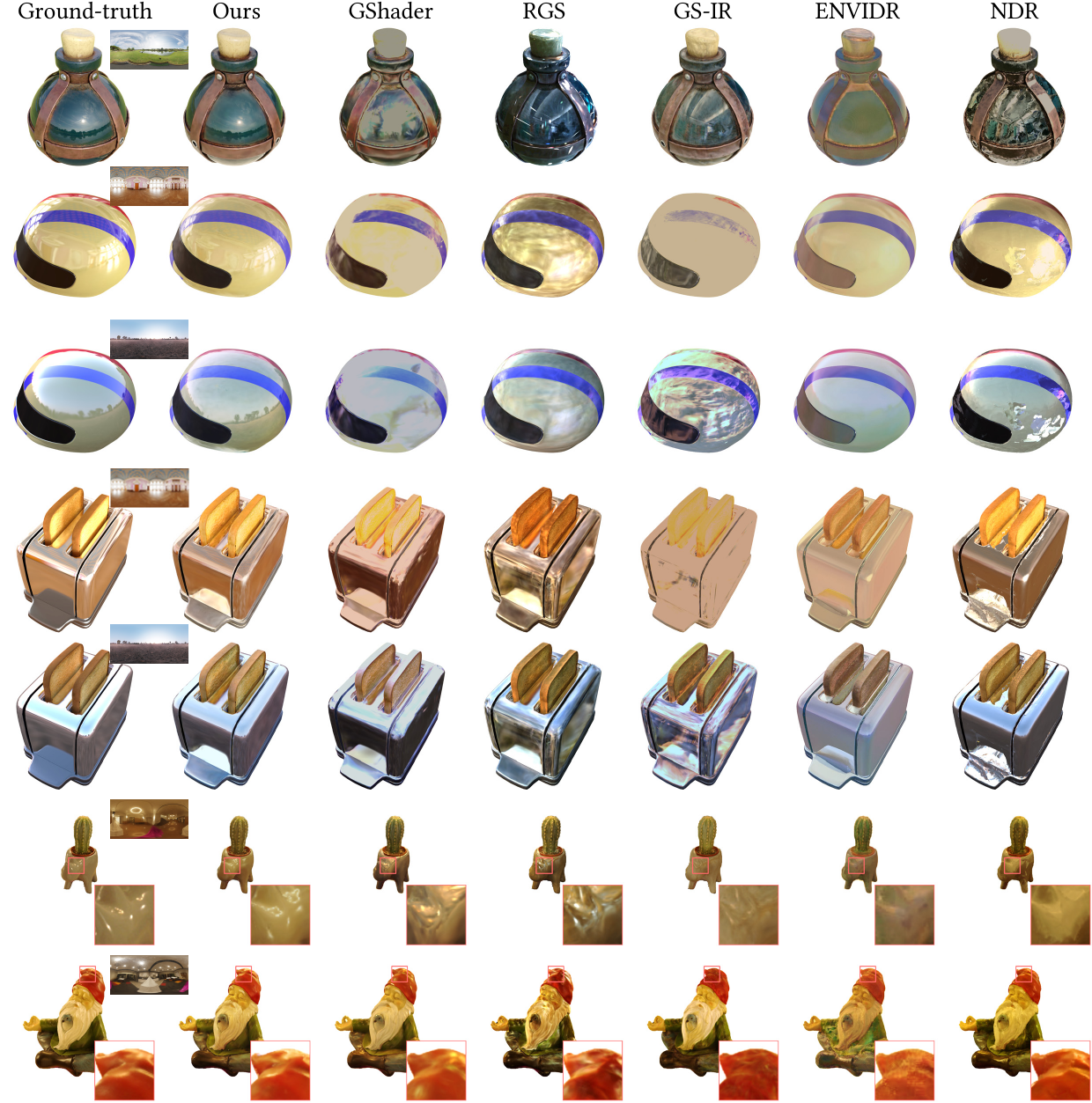}
\caption{Qualitative comparison of relighting results predicted by different methods. We scale the light for each method to cancel out inherent light-material ambiguity. From top to bottom: potion (from Glossy Synthetic~\cite{nero}), helmet and toaster (from Shiny Blender~\cite{ref_nerf}. Each takes up two rows), cactus and gnome (from Stanford ORB~\cite{stanfordorb}).  \label{fig:relight_cp}}
\end{figure*}

\begin{table*}[t]
\caption{Quantitative comparison of relighting results using PSNR, SSIM, and LPIPS metrics. Results are averaged over 20 different viewpoints with four different environment maps on Shiny Blender and Glossy Synthetic datasets. For the Stanford ORB dataset, relighting results are evaluated on the provided 20 image-envmap pairs. The environment maps are scaled to offset the inherent material-light ambiguity.\label{tab:relight_cp}}
\tabcolsep=0.21cm
\renewcommand\arraystretch{1.2}
\input{tabs/relight_cp}
\end{table*}

\begin{table*}[t]
\caption{Averaged training time and rendering frame rates. Time for baking visibility of GS-IR and our method are 7 and 8 minutes, respectively. \ye{The FPS of NDRMC~\cite{ndrmc} is evaluated by Blender Cycles with 64 samples per pixel (spp) and denoising. }\label{tab:performance}}
\tabcolsep=0.23cm
\renewcommand\arraystretch{1.2}
\input{tabs/efficiency_cp}
\end{table*}

\begin{figure}[t]
\centering
\includegraphics[width=1.0\columnwidth]{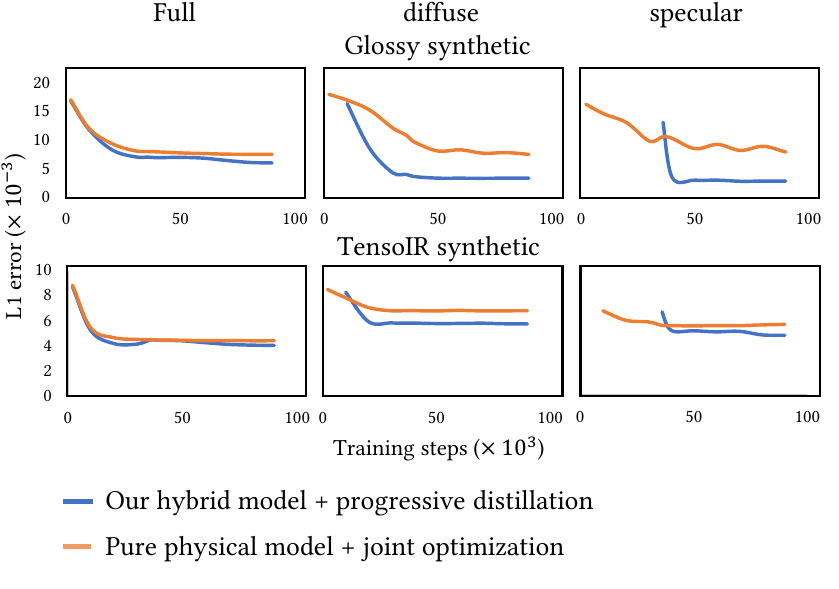}
\caption{\ye{L1 errors in the full rendering, diffuse and specular components as the training step increases. The first row is evaluated on Glossy synthetic dataset~\cite{nero} and the second row is evaluated on TensoIR synthetic dataset~\cite{jin2023tensoir}.}  \label{fig:loss}}
\end{figure}

We conduct comprehensive experiments on a workstation with an i7-13700KF CPU, 32GB memory and an NVIDIA RTX 4090 GPU, to demonstrate the effectiveness and efficiency of our approach. We also perform ablation studies to validate our core progressive distillation design.

\paragraph{Dataset.} We conduct evaluation on several datasets encompassing scenes with various materials, including \ye{three} synthetic datasets, Shiny Blender~\cite{ref_nerf}, Glossy Synthetic~\cite{nero} and TensoIR Synthetic~\cite{jin2023tensoir}, and one real captured dataset, Stanford ORB~\cite{stanfordorb}.
\ye{Shiny Blender and Glossy Synthetic mainly contains objects with low roughness while TensoIR Synthetic and Stanford ORB contains diffuse-dominant objects. All four datasets provide per-image object masks.}

\paragraph{Implementation details.} Before the specular distillation stage, we initialize the roughness $r$ and all channels of albedo $\boldsymbol{c}$ to 0.99. We apply a sigmoid activation to material properties $\boldsymbol{c}$, $r$ and $m$ to restrict their range to between 0 and 1. We apply an exponential activation to the light environment map to better reflect its HDR (High Dynamic Range) nature. $\mathrm{I_{phy}}$ is tone-mapped by the ACES Filmic tone mapping curve~\cite{arrighetti2017academy} and converted to the sRGB color space before image loss computation. The normal of each Gaussian is defined as its shortest axis facing the camera, which follows 3DGS-DR~\cite{our2024sig3dgsdf}. The environment light is represented as a $128\times128\times6$ cube map. The full rendering is based on a deferred shading pipeline, including two passes. In the first pass, maps of $\boldsymbol{c}$, $r$, $m$, $\mathrm{n}$, $d$, $\alpha$, and $\mathrm{I_{raw}}$ are splatted. Here $d$ is the depth map, computed by splatting a per-Gaussian depth, defined as the distance from its center to the camera. The positions used for querying the visibility term for each pixel are computed from the blended $d$. In the second pass, the output radiance is computed from the aforementioned maps and pre-filtered environment light as described in \secref{sec:model}.

\paragraph{Baselines and metrics.} We compare our method against the following baselines: \textbf{3DGS}: original 3D Gaussian Splatting~\cite{3DGS}; \textbf{GShader}~\cite{jiang2023gaussianshader}: a method that shades each Gaussian with a reflection-aware shader. \textbf{RGS}~\cite{gao2023relightable}: a method that performs ray tracing to render Gaussians. \textbf{GS-IR}~\cite{liang2024gsir}: a method that uses deferred shading and physically-based rendering model similar to ours. \textbf{3DGS-DR}~\cite{our2024sig3dgsdf}: a method focusing on mirror reflection rendering with 3DGS. \textbf{NDR}~\cite{nvidiffrec}: a method that extracts meshes from optimized SDFs (Signed Distance Functions) and render meshes using image based lighting. \ye{\textbf{NDRMC}~\cite{ndrmc}: a method that extends NDR by replacing the image based renderer with Monte Carlo renderer and using a differentiable denoiser to reduce variance.} \textbf{ENVIDR}~\cite{envidr}: a SDF-based method using neural rendering for inverse rendering. \ye{\textbf{TensoIR}~\cite{jin2023tensoir}: a NeRF-based method that using MLPs to cache visibility and indirect lighting.} \textbf{Ref-NeRF}~\cite{ref_nerf}: a NeRF-based method that improves the quality of novel view synthesis of reflective objects.

For the task of novel view synthesis, we compare with all the above-mentioned methods. For relighting tasks, 3DGS, 3DGS-DR and Ref-NeRF lack or only have limited relighting capability, so we only compare with \ye{GShader, GS-IR, RGS, NDR, NDRMC, ENVIDR and TensoIR}. We present quantitative results measured with three standard metrics: PSNR, SSIM and LPIPS. \ye{The evaluation results of NDRMC~\cite{ndrmc} are rendered using 2000 samples per pixel (spp) for novel views and 64 spp with denoiser for relighting, consistent with their paper.}

We note that previous works~\cite{physg, stanfordorb} align each predicted image to ground-truth via channel-wise scale factors for relighting comparison, which aims to eliminate the inherent scale ambiguity in inverse rendering problems. This scheme may lead to different scale factors for different views of the same scene, which is unreasonable and impractical in real rendering applications. A more reasonable way is to calculate and use the same scale factor for the same scene. 
Therefore, we propose to align the ground-truth environment map to the predicted environment map via channel-wise scale factors, and use these factors to scale the new environment map for relighting. All quantitative and qualitative results related to relighting below are based on this adjustment. We also report quantitative results of relighting using different scaling schemes in the supplementary material. In addition, we use Mean Angular Error in degrees (MAE$^\circ$) to evaluate the accuracy of normal reconstruction. We evaluate environment map reconstruction accuracy with LPIPS to mitigate the inherent ambiguity.

\subsection{Comparisons with baselines}

\paragraph{Novel view synthesis.} \tabref{tab:novel_view_cp} presents the quantitative comparison results on three datasets. Our method demonstrates a clear advantage in terms of image quality on synthetic datasets and also shows comparable results on real datasets. For diffuse-dominant scenes, the differences in image quality among various methods are quite small. The visual comparisons on synthetic and real datasets are shown in \figref{fig:novel_view}, and the supplementary material provides results on more scenes \ye{(mainly diffuse objects)}. 


The results from 3DGS-DR~\cite{our2024sig3dgsdf} and ENVIDR~\cite{envidr} demonstrate good quality in most scenes. However, 3DGS-DR only models mirror reflection and its quality decreases when a scene contains a wide range of roughness, such as the excessively sharp reflections in the bands of \textit{potion}. ENVIDR~\cite{envidr} is based on SDFs and can over-smooth surface normal, as shown in \textit{luyu}. It can also produce regularly-patterned noise as in \textit{toaster}. NDR~\cite{nvidiffrec} may generate incorrect geometry, leading to incorrect reflections, as in \textit{car}. It also performs poorly for specular objects with high diffuse albedo, as in \textit{cactus}. GShader~\cite{jiang2023gaussianshader} and RGS~\cite{gao2023relightable}, both based on per-Gaussian shading, tend to blur specular reflections. Our method produces satisfactory results for objects with various materials and stays closer to the ground truth.

\paragraph{Decomposition.} We visualize and compare the reconstruction results of normal, albedo, and lighting, and showcase our fully decomposition results.

Correct normals are crucial for relighting. We show the qualitative comparison in~\figref{fig:normal_cp} and dataset-averaged mean angular error of estimated normal on the Shiny Blender Dataset in~\tabref{tab:normal_env_cp}. The results from 3DGS-DR~\cite{our2024sig3dgsdf} show some noise in \textit{teapot}. ENVIDR~\cite{envidr} produces good normals, but over-smooths geometry details at normal discontinuities due to its smoothness SDF prior. This manifested as bumps in the concave parts at the top of \textit{potion}, just below the bottle neck. Other methods generally produced noisy normals, especially for reflective surfaces. Our method can achieve normal reconstruction quality comparable with ENVIDR, while retaining sharp boundaries.

Albedo, by definition, is the base color of an object and should not contain any reflection. Due to differences in rendering models among different methods, as well as the inherent ambiguity between light and material, we only conduct qualitative comparison in~\figref{fig:albedo_cp}. We mainly focus on whether the fitted albedo has reflections removed, rather than how closely it matches the reference image. Based on this criterion, ENVIDR~\cite{envidr}, NDR~\cite{nvidiffrec}, and our method all produce reasonable albedo maps, while other methods fail to disambiguate albedo from reflections, especially on smooth surfaces.

Correctly separating lighting is essential for distilling the correct physical model. The qualitative and quantitative comparison results are shown in~\figref{fig:envlight_cp} and~\tabref{tab:normal_env_cp}.  To compensate for the fundamental light-albedo ambiguity, we equalize the total energy of each environment map pair before comparison, and use the less scaling-sensitive LPIPS metric for evaluation. As shown in~\figref{fig:envlight_cp}, \textit{helmet}, \textit{teapot} and \textit{toaster} exhibit distinctive specular reflections, and \textit{gnome} is diffuse dominant. 3DGS-DR~\cite{our2024sig3dgsdf} predicts accurate lights for specular objects, but fails in \textit{gnome} because the method is mainly designed for mirror reflection. GShader~\cite{jiang2023gaussianshader} can only reconstruct a few texels for each Gaussian, leading to a noisy image with many holes. SDF (Signed Distance Function) based methods (ENVIDR~\cite{envidr} and NDR~\cite{nvidiffrec}), struggle with light decomposition accuracy, leading to messy patterns in the lighting estimation. Our method reconstructs environment maps with almost full directional coverage for reflective objects, and also accurately locates the primary light sources for diffuse-dominant scenes.

We show more decomposition results in~\figref{fig:decomp}. Our method accurately identifies surfaces with different roughness as well as metallic and non-metallic reflections. The distillation progress map $\alpha$ is close to 1 in most pixels, but slightly lower in pixels where the physical model does not fit well, such as between the slices of bread in \textit{toaster} and the spherical handle of \textit{tbell}. The raw radiance $\mathrm{I_{raw}}$ compensates for indirect light, which improves the quality in novel view synthesis.

\paragraph{Relighting.} Relighting is our primary goal. Without resolving the ambiguities between material and lighting or reconstructing accurate geometry, relighting will exhibit significant errors. We show the quantitative and qualitative results in~\tabref{tab:relight_cp} and~\figref{fig:relight_cp}. Our method demonstrates a significant advantage across all datasets. The results from GShader~\cite{jiang2023gaussianshader}, RGS~\cite{gao2023relightable}, and GS-IR~\cite{liang2024gsir} differ significantly from ground truth due to their inability to produce correct geometry and resolve the ambiguity between albedo and reflection. As shown in \textit{potion} of~\figref{fig:relight_cp}, these methods erronically retain reflections from the original lighting in relighted images. For diffuse-dominant objects, the ambiguity of geometry-material-lighting is smaller, making it a simpler task. As demonstrated in \textit{gnome}, almost all methods produce plausible results. The results from NDR~\cite{nvidiffrec} show the impact of unstable geometry on relighting. As shown in \textit{helmet}, NDR has demonstrated mostly-correct reflections except for where the estimated geometry ends up bumpy. ENVIDR~\cite{envidr} shows robust relighting results. However, due to its use of MLPs for lighting representation, one has to synthesize a considerable amount of training data for each relighting environment. The overall process consumes more than an hour per environment map. Additionally, its reflections tend to miss details. Thanks to our rendering model and carefully designed training process, we are able to resolve most of the ambiguities and demonstrate high-quality relighting results.

\paragraph{Efficiency.} \tabref{tab:performance} lists the dataset-averaged training time and rendering frame rate of all tested methods. Frame rate values are computed as the reciprocal of averaged frame render time to reduce the impact of unfairly large numbers on simple scenes. Additionally, we list the consumed time for GS-IR~\cite{liang2024gsir} and our method to bake visibility (7 minutes for GS-IR and 8 minutes for ours) in the caption. NeRF-based methods, including Ref-NeRF~\cite{ref_nerf}, TensoIR~\cite{jin2023tensoir} and ENVIDR~\cite{envidr} require more than one hour training time, and do not support real-time rendering. Gaussian splatting based methods, including RGS~\cite{gao2023relightable}, GShader~\cite{jiang2023gaussianshader}, GS-IR~\cite{liang2024gsir}, 3DGS-DR~\cite{our2024sig3dgsdf}, and vanilla 3DGS~\cite{3DGS} can be trained to convergence within one hour, and are able to render with high frame rates except for RGS, which is based on the more expensive ray tracing. \ye{Similarly, NDRMC~\cite{ndrmc} is based on differentiable Monte Carlo renderer with denoiser. Although accurate path tracing may achieve higher quality, it comes at the cost of longer training time and reduced real-time performance.} Our method requires moderate training time and shows high real-time performance among relighting-supporting methods. \ye{$\mathrm{I_{raw}}$ compensates unmodeled light paths, enabling our rendering model based on image based lighting to achieve quality comparable to path tracing methods on novel view synthesis.}

\subsection{Ablation Study} 

\begin{table}
\caption{Ablation study. Dataset-averaged image quality with selectively disabled algorithm components. \label{tab:ablation}}
\tabcolsep=0.14cm
\renewcommand\arraystretch{1.2}
\input{tabs/ablation}
\end{table}

\begin{figure}[t]
\centering
\includegraphics[width=1.0\columnwidth]{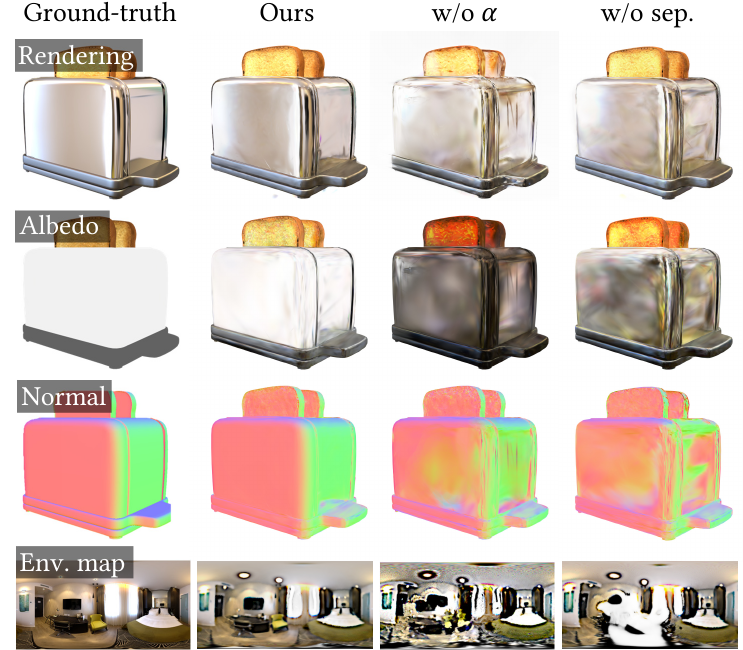}
\caption{Ablation studies. w/o $\alpha$ means directly using $\mathrm{I_{raw}}+\mathrm{I_{phy}}$ as the rendering model. w/o sep. means no separation of the specular and diffuse distillation.  \label{fig:ablation}}
\end{figure}

\begin{figure}[t]
\centering
\includegraphics[width=1.0\columnwidth]{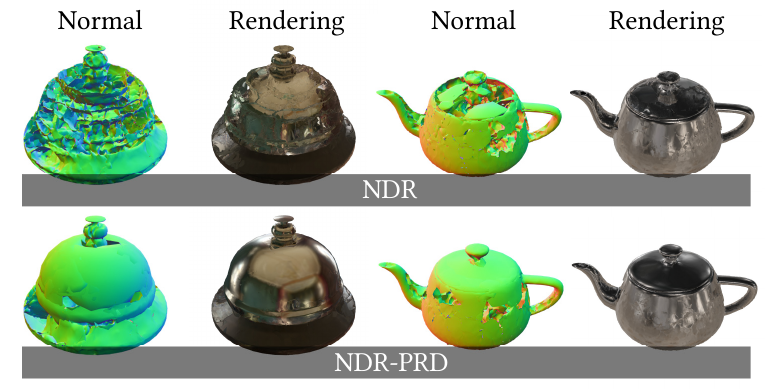}
\caption{Qualitative comparison of the original NDR~\cite{nvidiffrec} and modified NDR incorporating progressive radiance distillation (denoted as NDR-PRD). Our carefully designed optimization contributes to more accurate normal reconstruction for reflective objects.   \label{fig:ndr_cp}}
\end{figure}

\begin{table}[t]
\caption{Quantitative comparison of the original NDR~\cite{nvidiffrec} and modified NDR (denoted as NDR-PRD) that uses our optimization strategy as described in~\secref{sec:train}. Results are evaluated with PSNR metric.\label{tab:ndr_cp}}
\tabcolsep=0.13cm
\renewcommand\arraystretch{1.2}
\input{tabs/ndr_cp}
\end{table}

We conduct ablation studies to validate our core designs: the use of the distillation progress map and the separation of specular and diffuse distillation. The quantitative and qualitative results are shown in~\tabref{tab:ablation} and~\figref{fig:ablation}, which demonstrate the significance of our contribution.

\ye{\paragraph{Joint optimization from scratch vs. progressive distillation.}  To intuitively demonstrate the benefits of progressive distillation, we analyze the distillation process through visualizing the reconstruction quality of the diffuse/specular components and the full rendering during optimization. As shown in~\figref{fig:loss}, compared to jointly optimizing a pure physical model ($\mathrm{I}(\x,\wo)=\mathrm{I_{phy}}(\x,\wo)$) from scratch (referred to as JNT method), our method has similar specular and full rendering errors in early iterations but converges fast with smaller errors, which demonstrates that our distillation strategy effectively reduce ambiguities between different components. Consequently, the error in diffuse/specular components decreases rapidly after a few iterations, rather than slowly decreasing with fluctuations as seen in the JNT method. Moreover, although the JNT method shows significantly larger errors in the diffuse/specular components compared to our method, the error in full rendering is not as pronounced. This indicates that the diffuse component is contaminated with specular reflections, which will lead to a decrease in relighting quality, similar to the \textit{potion} results of RGS and GS-IR in~\figref{fig:relight_cp}.}

\paragraph{Impact of the distillation progress map.} We carefully designed a training process without using the distillation progress map. Specifically, the training still begins with a pre-training stage, during which $\mathrm{I_{raw}}$ and basic geometry parameters are optimized. In the second stage (and also the subsequent stages), the output radiance is:
\begin{equation}
    \mathrm{I}(\x,\wo)=\mathrm{I_{phy}}(\x,\wo)+\mathrm{I_{raw}}(\x,\wo).
\end{equation}
Normal propagation and color sabotage are still performed in this stage to help geometry reconstruction. In the third stage, we use a loss term to encourage $\mathrm{I_{raw}}$ to be zero, pushing the model to be physical. In the fourth stage, we drop that loss term and decrease the learning rate of the $\mathrm{I_{raw}}$ to fine tune everything. We believe this experiment design maximizes the capability of novel view synthesis and relighting in the absence of an explicit distillation progress. We refer to this experiment design as \textit{w/o $\alpha$}.

Without the guidance of the distillation progress map, the complicated physical model and light need to be optimized from the messy starting normal, during which each pixel will back-propagate a considerable amount of gradient to lighting and material attributes. This greatly increases the risk of getting stuck in local optima. As illustrated in~\figref{fig:ablation}, the environment map ends up containing several copies of the blackboard and the chair, because the optimization tends to copy reflected objects to explain incorrect normal. Our method initializes the distillation progress map with small values, which diverts most of the bad early gradient to the raw radiance, alleviating its negative impact on parameters to be estimated. When the fitting loss becomes favorable for the physical model, the distillation progress value starts to rise, simultaneously suppressing the contribution of raw radiance, which further reduces ambiguity and promotes the raw-to-physical transition. With the guidance of the distillation progress map, optimization can robustly proceed until specular reflections are fully distilled from raw radiance at the end of the second stage. In the third diffuse stage, we just \emph{move} the remaining parts from being represented by raw radiance to being represented by diffuse reflection.

\paragraph{Impact of the separation of specular and diffuse distillation.} Here we only keep the non-diffuse Stages 1, 2 and 4 in the original design, and replace~\equref{equ:stage2} in Stage 2 with the full rendering model~\equref{equ:fullModel}. We refer to this experiment design as \textit{w/o sep.} As shown in~\figref{fig:ablation}, the albedo map ends up entangled with random fragments of reflection, and the environment map missed some details. Although specular reflection and diffuse reflection can be theoretically differentiated using their angular frequency in the view-domain, the diffuse term propagates significant gradient to light and geometry, unlike the relatively independent raw radiance. Therefore, reflections that cannot be explained by low frequency effects may get forced into the diffuse term through erroneous normals, albedo, and lighting adjustments, disrupting optimization. Our method distills specular reflection first, and the converged results serve as a soft pinning for the lighting and geometry parameters, thereby avoiding the risk of a misguided optimizer using the diffuse term to explain specular reflection.

\paragraph{Impact of $\mathcal{L}_\alpha$.} We use an extra loss term $\mathcal{L}_\alpha$ to promote physical modeling. As shown in~\tabref{tab:ablation}, the $\mathcal{L}_\alpha$ term has a negligible impact on novel view synthesis. Being introduced after a converged specular stage, it primarily drives the transition of low-frequency representations from raw radiance to diffuse. Without $\mathcal{L}_\alpha$, raw radiance lingers on diffuse objects yet cannot respond to changes in lighting, leading to a significant decline in relighting quality.

\paragraph{Impact of $\mathrm{I_{raw}}$.} \ye{We retain $\mathrm{I_{raw}}$ in the final model to compensate for unfactorizable light paths and brake the distillation progress when it would lead to view interpolation regressions, which preserves the novel view quality not inferior to vanilla 3DGS. $\mathrm{I_{raw}}$ can be regarded as a self-emissive term, which can be easily supported in off-the-shelf engines with shaders. Furthermore, as shown in~\tabref{tab:ablation}, it has low weights in general with negligible impacts on relighting plausibility.}

\subsection{Discussion and Limitations}
\label{sec:discuss}

\begin{figure}[t]
\centering
\includegraphics[width=1.0\columnwidth]{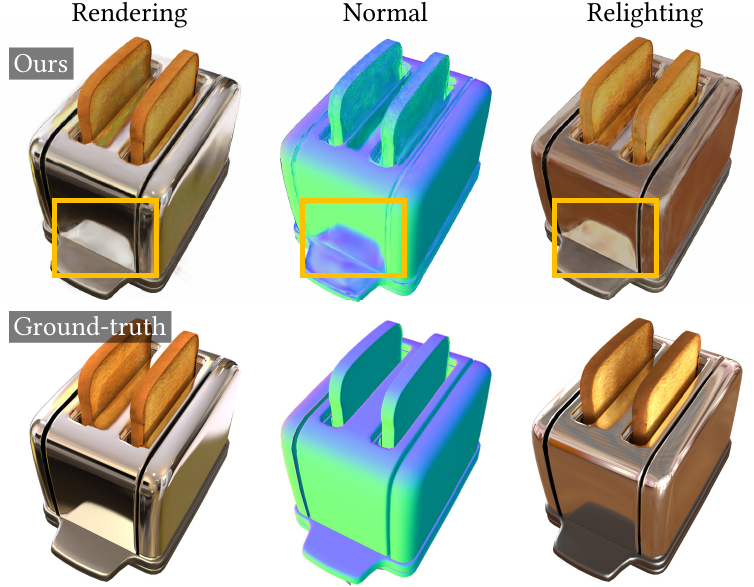}
\caption{Limitations. Inter-reflections mislead normal reconstruction, resulting in incorrect relighting.   \label{fig:limit}}
\end{figure}

Note that the idea of progressive radiance distillation is not limited to 3D Gaussian splatting, and it is possible to generalize it to NeRF representations. To explore this possibility, we adapted the rendering model and optimization strategy of NDR~\cite{nvidiffrec}, a representative mesh-based inverse rendering method, to incorporate our radiance distillation. NDR extracts meshes from SDF and uses the same PBR model as ours for inverse rendering. The output radiance of NDR is $\mathrm{I}(\x,\wo)=\mathrm{I_{AO}}\mathrm{I_{phy}}$ throughout its entire training process, where $\mathrm{I_{AO}}$ is a learned ambient occlusion texture. We remove $\mathrm{I_{AO}}$ and introduce $\mathrm{I_{raw}}$ and $\alpha$ as new learned textures, and employ \equref{equ:fullModel} as the rendering model. We modify NDR to follow the same four stages as our method, with the output radiance of each stage consistent with~\secref{sec:train}. We denote the modified NDR as NDR-PRD. The detailed implementation of NDR-PRD is described in the supplementary material.

The dataset-averaged PSNR metric of novel view synthesis and relighting are shown in~\tabref{tab:ndr_cp}. Quantitatively, NDR-PRD achieves slightly better results on the Shiny Blender and Glossy Synthetic dataset, but performs slightly worse on the Stanford ORB dataset. This could be attributed to the discontinuities in mesh based optimization, leading to a more challenging normal reconstruction, thus offsetting the benefits provided by our method. However, for a few scenes with specular reflections, NDR-PRD produces significantly better results, as shown in~\figref{fig:ndr_cp}. We provide more experimental results in the supplementary material.

The primary limitation of our method is the inability to handle indirect lighting explicitly, which could result in incorrect normals on surfaces with specular inter-reflections, and indirect light represented by raw radiance retaining the original lighting conditions when relighting, as shown in~\figref{fig:limit}. \ye{Besides, we use 3-order SH to compress the visibility at grid points, which is more suitable for soft shadows on diffuse surfaces. As shown in~\figref{fig:albedo_cp}, shadows may be baked into the albedo.} Performing ray tracing for inverse rendering like RGS~\cite{gao2023relightable} may solve these problems but significantly increase the rendering time. Another limitation arises from deferred shading, which affects the rendering quality of transparent or translucent objects like glass.

\section{Conclusion}

We have presented an inverse rendering method based on 3D Gaussian splatting, featuring a rendering model combining the radiance field rendering and physically-based rendering with a distillation progress map, and a stage-wise distillation strategy alleviating the inherent ambiguity of light-material decomposition. Extensive experiments demonstrate our competitive visual quality and real-time performance in both novel view synthesis and relighting. Our progressive radiance distillation also shows the potential for generalization to mesh-based inverse rendering. It will be interesting to fully explore the impact of combining our method with existing approaches.



{
\appendices

\section{BRDF model} For $\mathrm{I_{spec}}$, the specular component of the physically-based term $\mathrm{I_{phy}}$, we adopt the Cook-Torrance microfacet specular shading model:

\begin{equation}
    \mathrm{\rho}(\wi,\wo) = \frac{DGF}{4(\wo\cdot\mathbf{n})(\wi\cdot\mathbf{n})},
\end{equation}
where $D$, $G$ and $F$ are respectively the GGX~\cite{walter2007microfacet} normal distribution function (NDF), the geometric attenuation function and the approximated Fresnel term. Their specific expressions are as follows:

\begin{equation}
    F=F_0+(1-F_0)(1-(\mathbf{h} \cdot \wo))^5.
\end{equation}
\begin{equation}
    G(\mathbf{n}, \wo, \wi, k)=G_{\mathrm{sub }}(\mathbf{n}, \wo, k) G_{\mathrm{sub }}(\mathbf{n}, \wi, k),
\end{equation}
\begin{equation}
    G_{\mathrm {sub }}(\mathbf{n}, \mathbf{v}, k)=\frac{\mathbf{n} \cdot \mathbf{v}}{(\mathbf{n} \cdot \mathbf{v})(1-k)+k},
\end{equation}
\begin{equation}
    D(\mathbf{n}, \mathbf{h}, a)=\frac{a^2}{\pi((\mathbf{n} \cdot \mathbf{h})^2(a^2-1)+1)^2},
\end{equation}
where $\mathbf{h}$ is the half-way vector, a unit vector in the direction of $\wo+\wi$. $a=r^2$, $k=r^4/2$ are scalars derived from the roughness $r$. and $F_0$ is the basic reflection ratio:
\begin{equation}
    F_0 = m*\boldsymbol{c}+(1-m)*0.04,
    \label{equ:f0}
\end{equation}
where $m$ is the metallic map and $\boldsymbol{c}$ is the albedo.

We adopt the split-sum~\cite{hill2020physically} system to approximate the $\mathrm{I_{spec}}$ as:

\begin{equation}
    \mathrm{I_{spec}}(x,\wo) \approx \int_{\Omega} L(\wi) D(\mathbf{\wi},\mathbf{\wR},r^2) d \wi \cdot \int_{\Omega} \frac{D F G}{4(\wo \cdot \mathbf{n})} d \wi,
\end{equation}
where $\wR$ is the reflection direction $2(\wo \cdot \mathbf{n})\mathbf{n}-\wo$. The first part of the convolution is referred to as the pre-filtered environment maps, which are implemented as a mip-map pyramid, using lower resolution for higher roughness. The second part is the BRDF integration map, which is independent of lighting and can be precomputed into a lookup table. After precomputation, real-time physically-based rendering can be achieved, and the entire process is differentiable.

In the specular distillation stage, we fix $m=1$, which simplifies~\equref{equ:f0} to $F_0=\boldsymbol{c}$. Therefore, the albedo $\boldsymbol{c}$ at this stage actually serves as the basic reflection ratio, and it will be repurposed as the actual albedo in the diffuse distillation stage.

\begin{figure*}[t]
\centering
\includegraphics[width=1.0\linewidth]{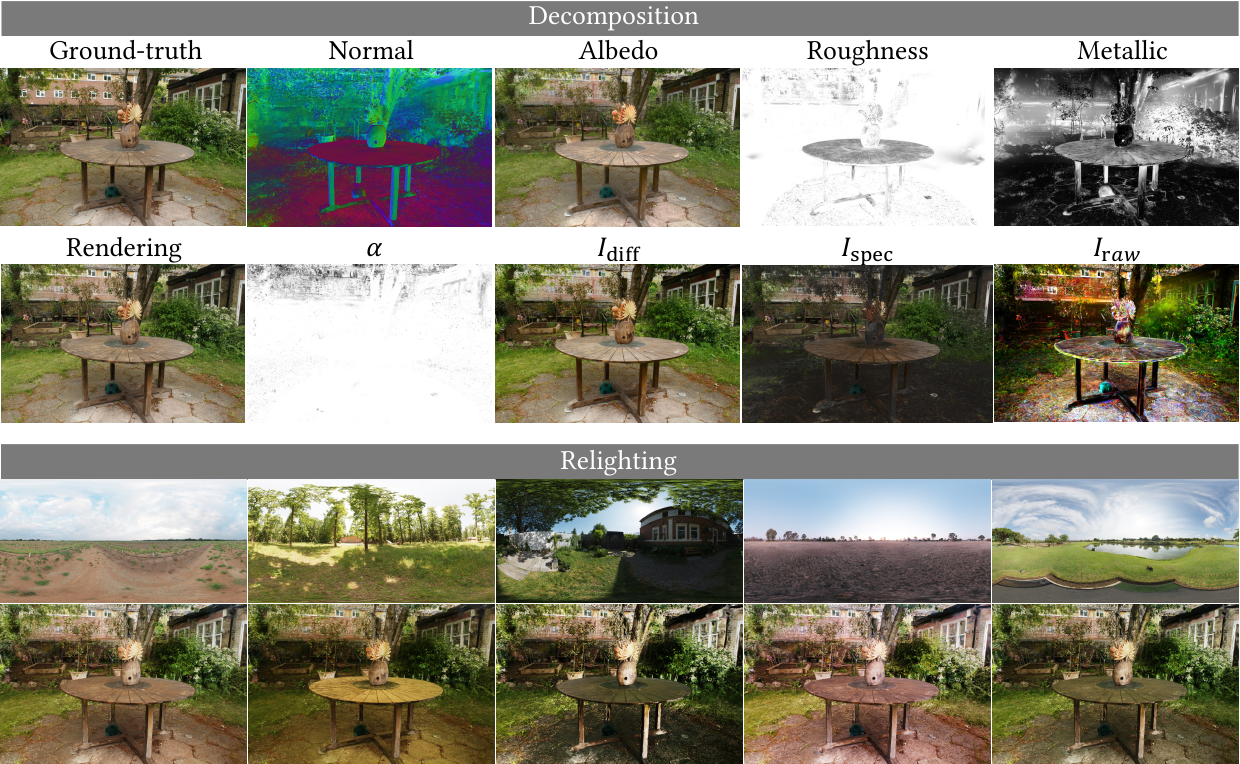}
\caption{Decomposition and relighting results of \textit{garden}. \label{fig:garden}}
\end{figure*}

For the diffuse part $\mathrm{I_{diff}}$, we implement the visibility term as a shadow field~\cite{zhou2005precomputed}. The final rendering step starts by projecting both the environment light and the visibility to order-3 SH basis:

\begin{equation}
    L(\wi)=\sum_{l=0}^2 \sum_{m=-l}^l l_{l}^m \mathrm{Y}_l^m(\wi), 
    \label{equ:L2sh}
\end{equation}
\begin{equation}
    V(x, \wi)=\sum_{l=0}^2 \sum_{m=-l}^l v_{l,x}^m \mathrm{Y}_l^m(\wi),
    \label{equ:V2sh}
\end{equation}
where $\mathrm{Y}_l^m(\wi)$ is the spherical harmonics basis, $l_{l}^m$ and $v_{l,x}^m$ are the corresponding coefficients. We perform SH triple product~\cite{sloan2008stupid} to project $V(x, \wi)L(\wi)$ back to the order-3 SH basis:
\begin{equation}
    V(x, \wi)L(\wi)=\sum_{l=0}^2 \sum_{m=-l}^l p_{l,x}^m \mathrm{Y}_l^m(\wi),
\end{equation}
where $p_{l,x}$ is the a tensor product of $l_{l}^m$, $v_{l,x}^m$ and the triple product constant tensor $C$. Dropping $x$ for clarity and merging $l,m$ pairs into linear indices $i$, $j$ and $k$, we get:
\begin{equation}
    p_i=\sum_j \sum_k C_{i j k} l_j v_k,
\end{equation}
\begin{equation}
    C_{i j k}=\int_{\Omega}\mathrm{Y}_i(\wi)\mathrm{Y}_j(\wi)\mathrm{Y}_k(\wi) d \wi.
\end{equation}

As explained by~\cite{ramamoorthi2001relationship}, the remaining cosine term $(\mathbf{n}\cdot\wi)$ and the hemi-spherical integration domain are azimuthally symmetric and the corresponding order-3 SH coefficients $\rho_l^m$ can be computed symbolically using Zonal Harmonics (ZH) rotation. The irradiance $\int_\Omega \mathrm{V}(\x,\wi)\mathrm{L}(\wi) (\mathbf{n}_x \cdot \wi) d \wi$ is thus a dot product between $\rho_l^m$ and $k_l^m$. In conclusion, we approximate $\mathrm{I_{diff}}$ as:
\begin{equation}
    \mathrm{I_{diff}}(x)\approx\frac{\boldsymbol{c}_x}{\pi} \sum_{l=0}^2 \sum_{m=-l}^l p_l^m \rho_l^m,
\end{equation}
or in the expanded triple-product form:
\begin{equation}
    \mathrm{I_{diff}}(x)\approx\frac{\boldsymbol{c}_x}{\pi} \sum_i \sum_j \sum_k C_{i j k} \rho_i l_j v_k,
\end{equation}


\section{Visibility baking}
To obtain the SH visibility term, we must first build a shadow field. We implement it as a regular grid ($120\times120\times120$ in our experiments) that fills the scene bounding box, where visibility SH coefficients are computed and saved on each grid point. Specifically, we first set all Gaussian colors to completely black and splat them into a white background from six cube face cameras centered at the grid point to form a cubemap. Then the rendered cubemaps are projected to the order-3 SH basis for coefficients. For any 3D point in the scene, its SH visibility can thus be queried from the grid using trilinear interpolation.

\begin{table*}[t]
\caption{Quantitative comparison of relighting results using PSNR, SSIM, and LPIPS metrics. Environment map scaling is disabled to better compare the absolute decomposition accuracy. \label{tab:relight_raw_cp}}
\tabcolsep=0.21cm
\renewcommand\arraystretch{1.2}
\input{tabs/relight_raw_cp}
\end{table*}

\begin{table*}[t]
\caption{Quantitative comparison of relighting results using PSNR, SSIM, and LPIPS metrics. Each predicted image is aligned to ground-truth via channel-wise scaling before computing metrics, as described in \cite{physg, stanfordorb}. \label{tab:relight_pim_cp}}
\tabcolsep=0.21cm
\renewcommand\arraystretch{1.2}
\input{tabs/relight_pim_cp}
\end{table*}

\section{Maskless real-world scenes}
Most inverse rendering methods depend on accurate per-object masks~\cite{physg,nerfactor,nvidiffrec}, whereas our approach only relies on masks in the diffuse distillation stage, in the $\mathcal{L}_\alpha$ term we introduced to promote physical modeling. However, there is an alternative design for $\mathcal{L}_\alpha$ that simply encourages the $\alpha$ of each Gaussian to approach 1:

\begin{equation}
    \mathcal{L}_\alpha^*=\sum_{i} \left|1-\alpha_i\right|,
    \label{equ:alpha_loss_new}
\end{equation}
where $i$ is the Gaussian index. We find that using $\mathcal{L}_\alpha^*$ might lead to slightly lower quality results for synthetic data (the average PSNR decreases by approximately 0.1 for novel view synthesis). However, $\mathcal{L}_\alpha^*$, combined with our rendering model, allows our method to be applied to any real-world scenes without relying on an object mask. We can also manually control which areas need to be fully decomposed and which areas retain the raw radiance rendering. Our rendering model can seamlessly concatenate these areas.

Take the real-world scene \textit{garden} provided by Mip-NeRF360~\cite{mip_nerf_360} as an example. The decomposition and relighting results are shown in~\figref{fig:garden}. We use a sphere domain to approximately cover the table and consider objects outside the sphere as background. In the diffuse distillation stage, we use~\equref{equ:alpha_loss_new} to promote the decomposition of Gaussians inside the sphere. In specular and diffuse distillation stage, we use the loss term $\mathcal{L}_{bkg}=\sum_{i}\alpha_i^2$, where $i$ loops over the indices of Gaussians outside the specified sphere, to encourage the background to fall back to raw radiance, which prevents it from interfering with environment light optimization. The results demonstrate the robustness of our method on in-the-wild data, making it more practical and user-friendly. Note that physically-based shading parameters of Gaussians outside the sphere domain generally become unconstrained and can be ignored. Another real-world scene we present is the \textit{sedan} from~\cite{ref_nerf}, used as our teaser figure. It employs the same processing method.

\section{Quantitative results of relighting with different scaling schemes}

We show the PSNR, SSIM and LPIPS metrics of relighting results without scaling the environment light in~\tabref{tab:relight_raw_cp}, which reflects the absolute decomposition accuracy to some extent. Our method demonstrates a clear advantage on the Shiny Blender and Glossy Synthetic datasets, and achieves results very close to the best. 


We also show relighting results with per-image scaling as described in~\cite{physg, stanfordorb} in~\tabref{tab:relight_pim_cp}. Our method outperforms alternatives regardless of the scaling schemes employed.

\begin{figure*}
\centering
\includegraphics[width=1.0\linewidth]{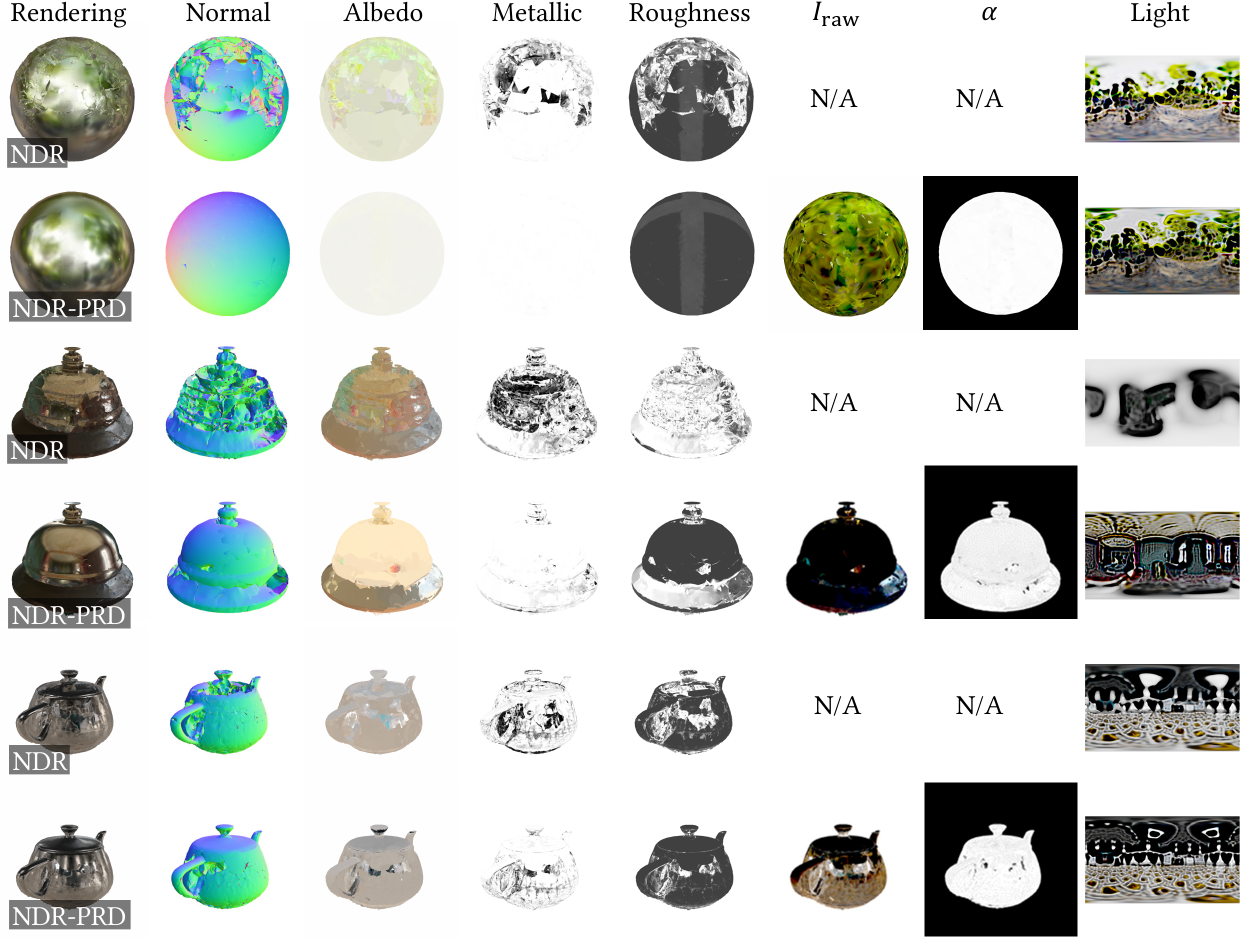}
\caption{Decomposition results of NDR~\cite{nvidiffrec} and NDR-PRD. \label{fig:ndrMore}}
\end{figure*}

\begin{figure}[t]
\centering
\includegraphics[width=1.0\columnwidth]{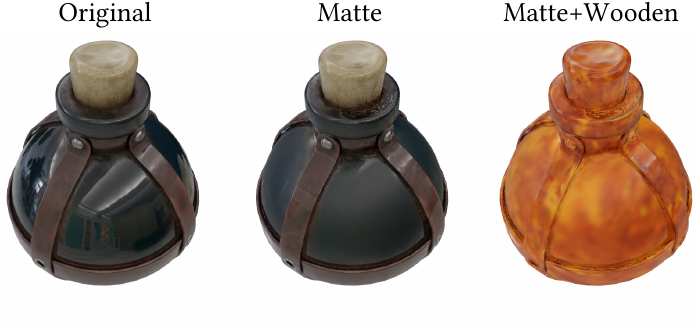}
\caption{Material editing. Left: the original \emph{potion} object. Middle: the same object with Matte material, simulated by flipping its roughness $r'=1-r$. Right: also replace its albedo with a volumetric wooden texture.\label{fig:matedit}}
\end{figure}

\section{Implementation details of NDR-PRD}

To apply progressive radiance distillation to NDR~\cite{nvidiffrec}, we use the same rendering model given by Eq.~(1) in Sec.~3.1 of the main paper, and develop a four-stage optimization procedure as described in Sec.~3.2. Any parameters unrelated to output radiance, including learning rate and loss weights, remain unchanged.
\paragraph{Pre-training} In this stage, meshes are extracted from the SDF and a position map is rendered from meshes. Following the original NDR~\cite{nvidiffrec}, we encode the positions and take them as the MLP input to predict the raw radiance $\mathrm{I_{raw}}$. The MLP architecture remains identical to the original NDR. We simply add three additional output channels for raw radiance. The same image loss is computed by comparing $\mathrm{I_{raw}}$ with ground-truth.
\paragraph{Specular distillation} The optimization in this stage is essentially the same as in the previous stage, except that the output radiance is replaced with the following:
\begin{equation}
    \mathrm{I}(\x,\wo)=\alpha_x\mathrm{I_{spec}}(x,\wo)+(1-\alpha_x)\mathrm{I_{raw}}(\x),
    \label{equ:stage2_NDR}
\end{equation}
where $\alpha$ is the progressive distillation map, which is also an additional MLP output. The PBR model used in NDR is identical to our method. The only difference is that NDR uses a $512\times512\times6$ cubemap, whereas we use $128\times128\times6$. Here we bumped our cubemap resolution of NDR-PRD to match NDR.

\paragraph{Diffuse distillation} Before starting this stage, following NDR, we extract meshes from the SDF and perform UV unwrapping. All attributes are then converted from MLP volume representations to optimizable 2D textures. The output radiance in this stage is the same as in our full rendering model. We use $\mathcal{L}_\alpha$ as an additional loss term to promote physically-based modeling. 
\paragraph{Joint refinement} In this stage, we drop $\mathcal{L}_\alpha$ but keep other regularization terms as the original NDR. The total number of training iterations for the four stages is equal to that of the original NDR.

\figref{fig:ndrMore} compares the decomposition results of NDR-PRD with the original NDR. Starting from a converged raw radiance, we are able to avoid early-stage local minima and our distillation appears significantly more sane for highly specular objects.

\section{Material editing}


In addition to relighting, our method also supports material editing. \figref{fig:matedit} illustrates one such example. Our high normal accuracy and clean distillation minimizes lingering fragments of the original texture, renders a convincing and thorough material change.

\section{More comparisons}

\figref{fig:sedan_cp} compares our method with GShader~\cite{jiang2023gaussianshader} on the \emph{sedan} scene. Our method produces smoother surface normal and managed to deduce the car material as almost completely specular. This allows us to reproduce the new-light coloration more faithfully. In contrast, GShader-relighted images exhibit visible remnants of the original reflection, with distinguishable tree-leaf patterns.


\ye{\figref{fig:tensoir_res} compares relighting results evaluated on TensoIR Synthetic dataset~\cite{jin2023tensoir}, which contains all diffuse-dominant objects. Compared to all baselines, our method achieves realistic highlights and soft shadows close to the ground-truth (e.g. shadows under the jaw of \textit{armadillo} on the second row, and highlights reflected by the plate on the sixth row). TensoIR~\cite{jin2023tensoir} achieves sharper shadows by caching the visibility using a MLP, but it fails to capture any highlights reflections and suffers from noticeable aliasing issues. NDRMC~\cite{ndrmc} achieves more accurate shadows and indirect lighting by using a differentiable Monte Carlo renderer, but produces fragmented meshes for slightly smoother surfaces (e.g. pot of \textit{ficus} on the third and forth row). Other methods fail to capture shading details and RGS~\cite{gao2023relightable} produces failure results under certain lighting.}

\begin{figure}[t]
\centering
\includegraphics[width=1.0\columnwidth]{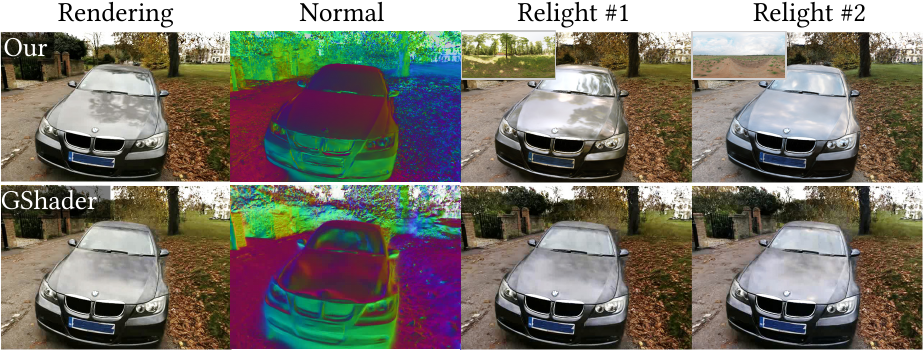}
\caption{Comparison with GShader~\cite{jiang2023gaussianshader} on the \emph{sedan} scene. \label{fig:sedan_cp}}
\end{figure}

\begin{figure*}
\centering
\includegraphics[width=1.0\linewidth]{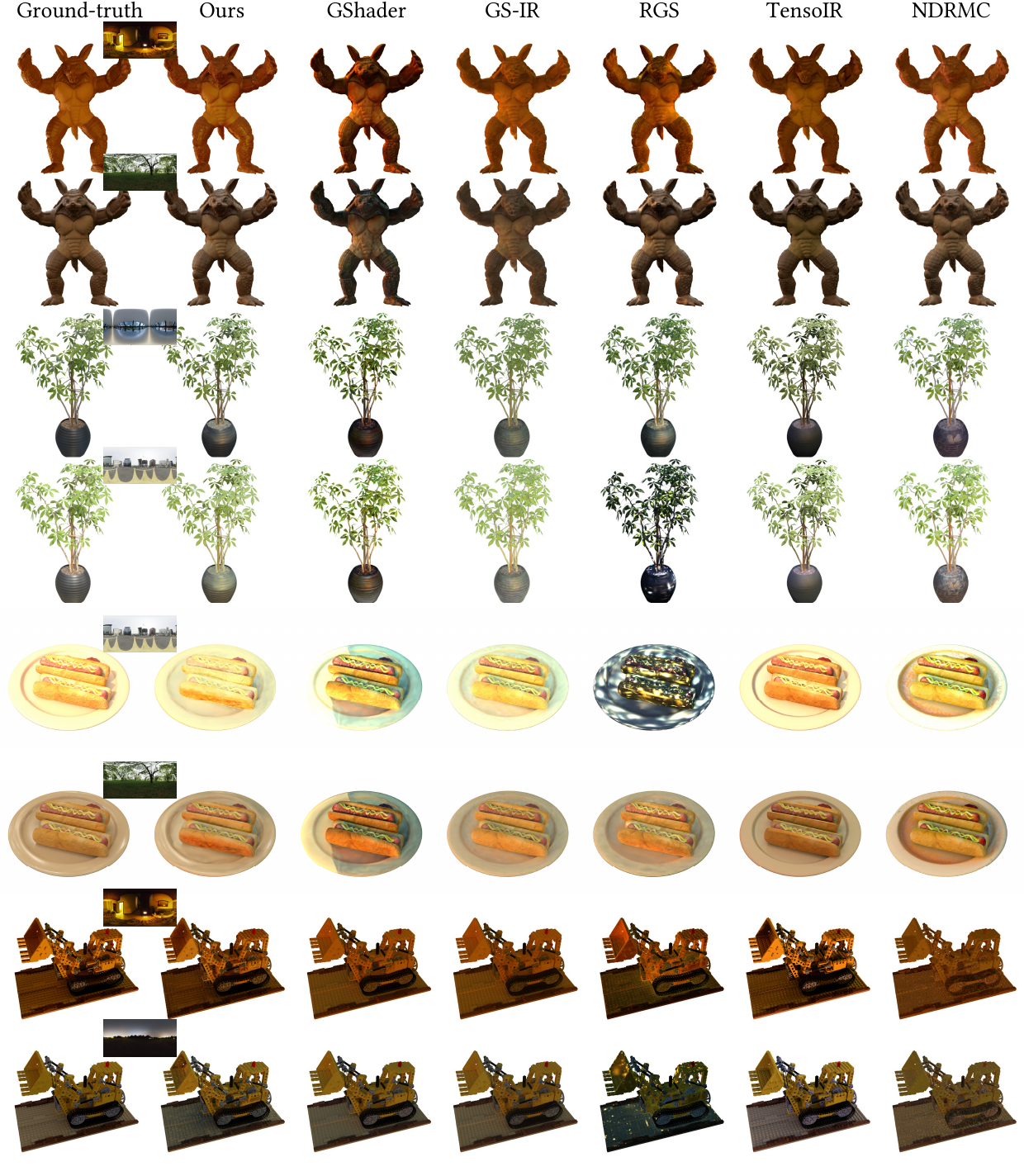}
\caption{More relighting results on TensoIR Synthetic dataset~\cite{jin2023tensoir}. All objects are diffuse-dominant. \label{fig:tensoir_res}}
\end{figure*}

\begin{figure*}
\centering
\includegraphics[width=1.0\linewidth]{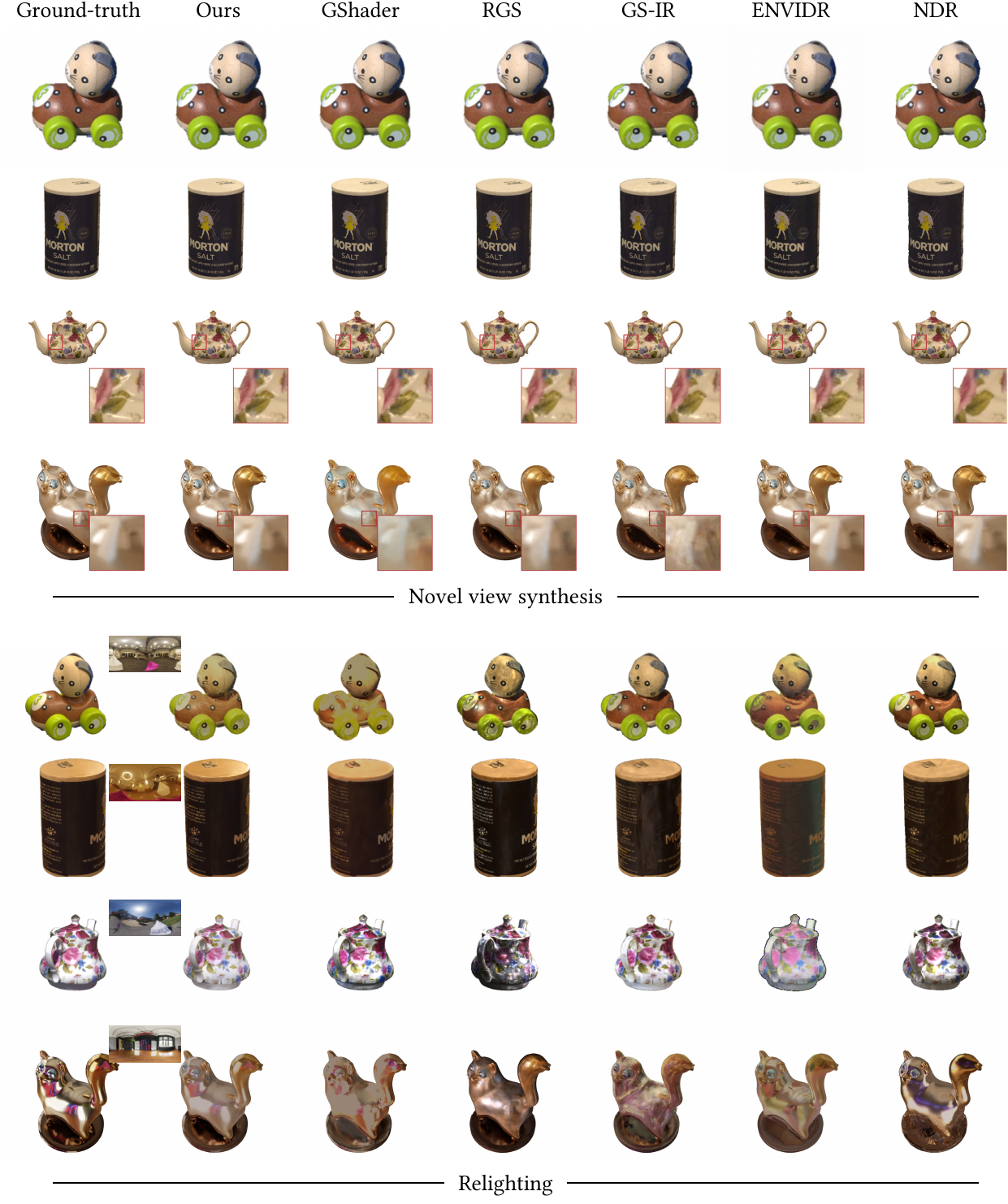}
\caption{More comparisons. Note that captions are placed below corresponding subfigures here. \label{fig:more_res}}
\end{figure*}

\figref{fig:more_res} compares novel view synthesis and relighting between our method and state-of-the-art techniques on more scenes.
}

\bibliographystyle{IEEEtran}
\bibliography{pdgs}


 





\end{document}

%% file: tabs/novel_view_cp.tex
\begin{tabular}{c|ccc|ccc|ccc|ccc}
\hline
\multirow{2}{*}{Methods} & \multicolumn{3}{c|}{Shiny Blender} & \multicolumn{3}{c|}{Glossy Synthetic} & \multicolumn{3}{c|}{TensoIR Synthetic} & \multicolumn{3}{c}{Stanford ORB} \\ \cline{2-13} 
                         & PSNR $\uparrow$       & SSIM $\uparrow$      & LPIPS $\downarrow$    & PSNR $\uparrow$        & SSIM $\uparrow$      & LPIPS $\downarrow$ 
                         & PSNR $\uparrow$     & SSIM $\uparrow$     & LPIPS $\downarrow$& PSNR $\uparrow$     & SSIM $\uparrow$     & LPIPS $\downarrow$   \\ \hline
Ref-NeRF                 & 33.13      & 0.961     & 0.080     & 27.50       & 0.927      & 0.100      & 38.42 & 0.974 & 0.026 & 35.43     & 0.983     & 0.028    \\
3DGS                     & 30.36      & 0.947     & 0.084     & 26.50       & 0.917      & 0.092      &39.07 & \cellcolor{c1}0.988 & \cellcolor{c1}0.016 & 36.34     & \cellcolor{c2}0.990     & \cellcolor{c2}0.013    \\
GShader                  & 31.97      & 0.958     & 0.067     & 27.54       & 0.922      & 0.087      & 38.78 & 0.980 & 0.021 & \cellcolor{c1}36.86     & \cellcolor{c1}0.992     & 0.153    \\
GS-IR                    & 26.63      & 0.878     & 0.150     & 23.79       & 0.827      & 0.161      & 34.98 & 0.962 & 0.043 & 34.21     & 0.968     & 0.038    \\
RGS                      & 29.19      & 0.940     & 0.100     & 25.64       & 0.911      & 0.099      & 39.01 & 0.981 & \cellcolor{c2}0.017 & 35.70     & 0.988     & 0.016    \\
NDR                      & 28.49      & 0.930     & 0.116     & 25.69       & 0.900      & 0.113      & 30.11 & 0.941 & 0.058 & 33.15     & 0.983     & 0.022    \\
NDRMC              &31.24 	&0.954 	&0.092 	&28.62 	&0.947 	&0.059 &30.03	&0.949	&0.083 &36.44 	&0.982 	&0.028   \\
TensoIR         &28.01 	&0.905 	&0.131 	&25.88 	&0.904 	&0.112 
 &35.09	&0.976	&0.04 &34.66 	&0.980 	&0.020  \\
ENVIDR                   & 33.46      & 0.968     & \cellcolor{c1}0.046     & 29.58       & 0.952      & \cellcolor{c2}0.056      & 30.76 & 0.943 & 0.051 & 34.78     & 0.981     & 0.022    \\
3DGS-DR                  & \cellcolor{c2}34.09      & \cellcolor{c2}0.971     & 0.057     & \cellcolor{c2}30.14       & \cellcolor{c2}0.953      & 0.058      & \cellcolor{c1}39.24 & \cellcolor{c1}0.988 & \cellcolor{c2}0.017 & 36.33     & \cellcolor{c2}0.990     & \cellcolor{c2}0.013    \\
Ours                     & \cellcolor{c1}34.43      & \cellcolor{c1}0.973     & \cellcolor{c2}0.054     & \cellcolor{c1}30.32       & \cellcolor{c1}0.958      & \cellcolor{c1}0.051      & \cellcolor{c2}39.11 & \cellcolor{c2}0.986 & \cellcolor{c2}0.017 &\cellcolor{c2}36.50     & \cellcolor{c2}0.990     & \cellcolor{c1}0.012    \\ \hline
\end{tabular}

%% file: tabs/normal_env_cp.tex
\begin{tabular}{c|cccccc}
\hline
      & GShader & GS-IR & NDR   & ENVIDR & 3DGS-DR & Ours  \\ \hline
MAE$^\circ$ $\downarrow$    & 23.31   & 31.24 & 17.02 & \cellcolor{c1}4.618  & 4.871   &  \cellcolor{c2}4.621 \\
LPIPS $\downarrow$  & 0.621   & 0.687 & 0.636 & 0.615  & \cellcolor{c1}0.511   & \cellcolor{c2}0.521 \\ \hline
\end{tabular}

%% file: tabs/relight_cp.tex
\begin{tabular}{c|ccc|ccc|ccc|ccc}
\hline
\multirow{2}{*}{Methods} & \multicolumn{3}{c|}{Shiny Blender} & \multicolumn{3}{c|}{Glossy Synthetic} & \multicolumn{3}{c|}{TensoIR Synthetic} & \multicolumn{3}{c}{Stanford ORB} \\ \cline{2-13} 
                         & PSNR $\uparrow$       & SSIM $\uparrow$      & LPIPS $\downarrow$    & PSNR $\uparrow$        & SSIM $\uparrow$      & LPIPS $\downarrow$ 
                         & PSNR $\uparrow$     & SSIM $\uparrow$     & LPIPS $\downarrow$& PSNR $\uparrow$     & SSIM $\uparrow$     & LPIPS $\downarrow$   \\ \hline
GShader                  & 20.26      & 0.881     & 0.130     & \cellcolor{c2}21.66       & 0.870      &\cellcolor{c2}0.120      & 25.61 & 0.927 & \cellcolor{c2}0.066  & 29.72     & 0.957     & 0.038    \\
GS-IR                    & 19.37      & 0.862     & 0.171     & 19.80       & 0.808      & 0.160     & 25.52 & 0.904 & 0.084  & 28.05     & 0.947     & 0.046    \\
RGS                      & 17.75      & 0.841     & 0.151     & 17.08       & 0.807      & 0.155     & 25.38 & 0.916 & 0.082  & 26.49     & 0.939     & 0.046    \\
NDR                & 21.81      & 0.891     & 0.133     & 19.82       & 0.826      & 0.153     & 26.59 & 0.906 & 0.091  & 29.07     & 0.953     & 0.041    \\
NDRMC              &\cellcolor{c2}21.92	&0.904 	&0.143 	&21.42 	&\cellcolor{c2}0.872 	&0.134 
&26.86	&0.907	&0.113  &\cellcolor{c2}30.88 	&\cellcolor{c1}0.962 	&0.040  \\
TensoIR              &19.01	&0.862 	&0.171 	&18.71 	&0.812 	&0.156 
&\cellcolor{c2}28.23	&\cellcolor{c2}0.938	&0.088  &30.01 	&0.957 	&\cellcolor{c2}0.036    \\
ENVIDR                   & 20.90      & \cellcolor{c2}0.905     & \cellcolor{c2}0.102     & 19.13       & 0.859      & 0.124     & 25.58 & 0.924 & 0.087  & 26.75     & 0.942     & 0.049    \\
Ours                     & \cellcolor{c1}23.68      & \cellcolor{c1}0.925     & \cellcolor{c1}0.083     & \cellcolor{c1}23.10       & \cellcolor{c1}0.895      & \cellcolor{c1}0.091      & \cellcolor{c1}29.69 & \cellcolor{c1}0.941 & \cellcolor{c1}0.057 & \cellcolor{c1}30.92     & \cellcolor{c2}0.960     & \cellcolor{c1}0.033    \\ \hline
\end{tabular}

%% file: tabs/efficiency_cp.tex
\begin{tabular}{c|ccccccccccc}
\hline
              & Ref-NeRF &NDRMC &TensoIR & ENVIDR & RGS   & NDR   & GShader & GS-IR    & 3DGS-DR & 3DGS & Ours     \\ \hline
Training time & 19h &4.2h      & 3.9h & 3.2h   & 41min & 78min & 60min   & 19+7 min & 13min   & 6min & 39+8 min \\
FPS           & 0.05 & 6 (64 spp) & 4     & 3      & 4     & 32    & 34      & 182      & 281     & 294  & 221      \\ \hline
\end{tabular}

%% file: tabs/ablation.tex
\begin{tabular}{c|ccc|ccc}
\hline
\multirow{2}{*}{Ablations} & \multicolumn{3}{c|}{Novel view synthesis} & \multicolumn{3}{c}{Relighting} \\
                           & PSNR $\uparrow$         & SSIM $\uparrow$        & LPIPS $\downarrow$      & PSNR $\uparrow$    & SSIM $\uparrow$    & LPIPS $\downarrow$   \\ \hline
w/o $\alpha$                  & 32.57        & 0.958        & 0.089       & 22.17    & 0.878    & 0.112    \\
w/o sep.                      & 33.04        & 0.965        & 0.064       & 23.09    & 0.909    & 0.095    \\
w/o $\mathcal{L}_\alpha$                  & \cellcolor{c1}34.45        & \cellcolor{c2} 0.972        & \cellcolor{c1}0.054       & 19.88    & 0.861    & 0.169    \\
w/o $\mathrm{I_{raw}}$                  & 32.97        & 0.964        & 0.066       & \cellcolor{c2}23.41    & \cellcolor{c2}0.912    & \cellcolor{c2}0.087    \\
Ours                       & \cellcolor{c2}34.43        & \cellcolor{c1}0.973        & \cellcolor{c1}0.054       & \cellcolor{c1}23.68    & \cellcolor{c1}0.925    & \cellcolor{c1}0.083    \\ \hline
\end{tabular}

%% file: tabs/ndr_cp.tex
\begin{tabular}{c|cc|cc|cc}
\hline
\multirow{2}{*}{} & \multicolumn{2}{c|}{Shiny Blender} & \multicolumn{2}{c|}{Glossy Synthetic} & \multicolumn{2}{c}{Stanford ORB} \\
                  & NDR              & NDR-PRD            & NDR               & NDR-PRD              & NDR             & NDR-PRD           \\ \hline
Novel view        & 28.49            & \cellcolor{c1}30.39           & 25.69             & \cellcolor{c1}26.49             & \cellcolor{c1}33.15           & 33.05          \\
Relighting        & 21.81            & \cellcolor{c1}22.09           & 19.82             & \cellcolor{c1}20.09             & \cellcolor{c1}29.07           & 28.43          \\ \hline
\end{tabular}

%% file: tabs/relight_raw_cp.tex
\begin{tabular}{c|ccc|ccc|ccc|ccc}
\hline
\multirow{2}{*}{Methods} & \multicolumn{3}{c|}{Shiny Blender} & \multicolumn{3}{c|}{Glossy Synthetic} & \multicolumn{3}{c|}{TensoIR Synthetic} & \multicolumn{3}{c}{Stanford ORB} \\ \cline{2-13} 
                         & PSNR $\uparrow$       & SSIM $\uparrow$      & LPIPS $\downarrow$    & PSNR $\uparrow$        & SSIM $\uparrow$      & LPIPS $\downarrow$ 
                         & PSNR $\uparrow$     & SSIM $\uparrow$     & LPIPS $\downarrow$& PSNR $\uparrow$     & SSIM $\uparrow$     & LPIPS $\downarrow$   \\ \hline
GShader                  & 16.22      & 0.868     & 0.152     & 16.32       & 0.838      & 0.135     & \cellcolor{c2}22.90 & \cellcolor{c1}0.922 & \cellcolor{c2}0.067  & 26.23     & \cellcolor{c1}0.952     & \cellcolor{c1}0.043    \\
GS-IR                    & 16.77      & 0.849     & 0.185     & \cellcolor{c2}19.36       & 0.802      & 0.162     & 22.52 & 0.888 & 0.094  & 25.01     & 0.935     & 0.050    \\
RGS                      & 14.08      & 0.831     & 0.160     & 14.70       & 0.823      & 0.152     & 20.88 & 0.906 & 0.094  & 22.01     & 0.903     & 0.060    \\
NDR                & \cellcolor{c2}21.73      & 0.893     & 0.137     & 16.97       & 0.809      & 0.162     & 18.68 & 0.872 & 0.113  & \cellcolor{c1}26.69     & 0.949     & \cellcolor{c2}0.044    \\
NDRMC                & 21.42      & 0.872     & 0.155     & 18.46       & 0.831      & 0.147     & 21.47 & 0.913 & 0.108  & \cellcolor{c2}26.64     & 0.941     & 0.046    \\
TensoIR                & 17.01      & 0.838     & 0.142     & 16.12       & 0.821      & 0.154     & \cellcolor{c1}22.97 & \cellcolor{c1}0.922 & \cellcolor{c2}0.067  & 26.22     & 0.947     & 0.049    \\
ENVIDR                   & 19.55      & \cellcolor{c2}0.898     & \cellcolor{c2}0.106     & 17.11       & \cellcolor{c2}0.844      & \cellcolor{c2}0.129      & 18.61 & 0.863 & 0.124 & 24.17     & 0.935     & 0.051    \\
Ours                     & \cellcolor{c1}22.48      & \cellcolor{c1}0.924     & \cellcolor{c1}0.086     & \cellcolor{c1}20.81       & \cellcolor{c1}0.892      & \cellcolor{c1}0.090      & 22.65 & \cellcolor{c2}0.921 & \cellcolor{c1}0.066 & 26.24     & \cellcolor{c2}0.950     & 0.045    \\ \hline
\end{tabular}

%% file: tabs/relight_pim_cp.tex
\begin{tabular}{c|ccc|ccc|ccc|ccc}
\hline
\multirow{2}{*}{Methods} & \multicolumn{3}{c|}{Shiny Blender} & \multicolumn{3}{c|}{Glossy Synthetic} & \multicolumn{3}{c|}{TensoIR Synthetic} & \multicolumn{3}{c}{Stanford ORB} \\ \cline{2-13} 
                         & PSNR $\uparrow$       & SSIM $\uparrow$      & LPIPS $\downarrow$    & PSNR $\uparrow$        & SSIM $\uparrow$      & LPIPS $\downarrow$ 
                         & PSNR $\uparrow$     & SSIM $\uparrow$     & LPIPS $\downarrow$& PSNR $\uparrow$     & SSIM $\uparrow$     & LPIPS $\downarrow$   \\ \hline
GShader                  & 20.99      & 0.889     & 0.125     & 21.80       & 0.870      & \cellcolor{c2}0.120     & 26.86 & 0.930 & \cellcolor{c2}0.063 & 30.32     & 0.958     & \cellcolor{c2}0.037    \\
GS-IR                    & 20.43      & 0.865     & 0.166     & 20.17       & 0.813      & 0.157     & 26.24 & 0.907 & 0.083 & 29.02     & 0.949     & 0.045    \\
RGS                      & 18.61      & 0.847     & 0.148     & 17.39       & 0.809      & 0.151    & 26.41 & 0.919 & 0.080  & 27.32     & 0.941     & 0.045    \\
NDR                & 22.65      & 0.896     & 0.131     & 20.02       & 0.827      & 0.153    & 27.10 & 0.907 & 0.091  & 29.44     & 0.953     & 0.040    \\
NDRMC              &\cellcolor{c2}23.91	&\cellcolor{c2}0.911 	&0.151 	&\cellcolor{c2}22.13 	&\cellcolor{c2}0.879 	&0.126 
&27.81	&0.907	&0.110  &\cellcolor{c1}31.52 	&\cellcolor{c1}0.967 	&0.039  \\
TensoIR              &19.17	&0.887 	&0.150 	&18.97 	&0.824 	&0.143 
&\cellcolor{c2}28.58	&\cellcolor{c1}0.944	&0.081  &30.60 	&0.960 	&\cellcolor{c1}0.033    \\
ENVIDR                   & 21.40      & 0.906     & \cellcolor{c2}0.101     & 19.27       & 0.860      & 0.124 & 26.21 & 0.911 & 0.114     & 26.64     & 0.942     & 0.048    \\
Ours                     & \cellcolor{c1}24.77      & \cellcolor{c1}0.928     & \cellcolor{c1}0.079     & \cellcolor{c1}23.30       & \cellcolor{c1}0.895      &\cellcolor{c1} 0.091      & \cellcolor{c1}30.29 & \cellcolor{c2}0.942 & \cellcolor{c1}0.056 & \cellcolor{c2}31.32     & \cellcolor{c2}0.961     & \cellcolor{c1}0.033    \\ \hline
\end{tabular}

%% file: pdgs_full.bbl
\begin{thebibliography}{10}
\providecommand{\url}[1]{#1}
\csname url@samestyle\endcsname
\providecommand{\newblock}{\relax}
\providecommand{\bibinfo}[2]{#2}
\providecommand{\BIBentrySTDinterwordspacing}{\spaceskip=0pt\relax}
\providecommand{\BIBentryALTinterwordstretchfactor}{4}
\providecommand{\BIBentryALTinterwordspacing}{\spaceskip=\fontdimen2\font plus
\BIBentryALTinterwordstretchfactor\fontdimen3\font minus \fontdimen4\font\relax}
\providecommand{\BIBforeignlanguage}[2]{{%
\expandafter\ifx\csname l@#1\endcsname\relax
\typeout{** WARNING: IEEEtran.bst: No hyphenation pattern has been}%
\typeout{** loaded for the language `#1'. Using the pattern for}%
\typeout{** the default language instead.}%
\else
\language=\csname l@#1\endcsname
\fi
#2}}
\providecommand{\BIBdecl}{\relax}
\BIBdecl

\bibitem{ref_nerf}
D.~Verbin, P.~Hedman, B.~Mildenhall, T.~Zickler, J.~T. Barron, and P.~P. Srinivasan, ``Ref-nerf: Structured view-dependent appearance for neural radiance fields,'' in \emph{Proceedings of the IEEE/CVF CVPR}.\hskip 1em plus 0.5em minus 0.4em\relax IEEE, 2022, pp. 5481--5490.

\bibitem{render_equation}
J.~T. Kajiya, ``The rendering equation,'' in \emph{Proceedings of the 13th annual conference on Computer graphics and interactive techniques}, 1986, pp. 143--150.

\bibitem{nerf}
B.~Mildenhall, P.~P. Srinivasan, M.~Tancik, J.~T. Barron, R.~Ramamoorthi, and R.~Ng, ``Nerf: Representing scenes as neural radiance fields for view synthesis,'' in \emph{Proceedings of ECCV}, 2020, pp. 405--421.

\bibitem{3DGS}
B.~Kerbl, G.~Kopanas, T.~Leimk{\"u}hler, and G.~Drettakis, ``3d gaussian splatting for real-time radiance field rendering,'' \emph{ACM Trans. Graph.}, vol.~42, no.~4, July 2023.

\bibitem{nerfactor}
X.~Zhang, P.~P. Srinivasan, B.~Deng, P.~Debevec, W.~T. Freeman, and J.~T. Barron, ``Nerfactor: Neural factorization of shape and reflectance under an unknown illumination,'' \emph{ACM Trans. Graph.}, vol.~40, no.~6, 2021.

\bibitem{nerd}
M.~Boss, R.~Braun, V.~Jampani, J.~T. Barron, C.~Liu, and H.~Lensch, ``Nerd: Neural reflectance decomposition from image collections,'' in \emph{Proceedings of the IEEE/CVF CVPR}, 2021, pp. 12\,684--12\,694.

\bibitem{nero}
Y.~Liu, P.~Wang, C.~Lin, X.~Long, J.~Wang, L.~Liu, T.~Komura, and W.~Wang, ``Nero: Neural geometry and brdf reconstruction of reflective objects from multiview images,'' \emph{ACM Trans. Graph.}, vol.~42, no.~4, jul 2023.

\bibitem{jiang2023gaussianshader}
Y.~Jiang, J.~Tu, Y.~Liu, X.~Gao, X.~Long, W.~Wang, and Y.~Ma, ``Gaussianshader: 3d gaussian splatting with shading functions for reflective surfaces,'' 2023.

\bibitem{liang2024gsir}
Z.~Liang, Q.~Zhang, Y.~Feng, Y.~Shan, and K.~Jia, ``Gs-ir: 3d gaussian splatting for inverse rendering,'' 2024.

\bibitem{GIR}
Y.~Shi, Y.~Wu, C.~Wu, X.~Liu, C.~Zhao, H.~Feng, J.~Liu, L.~Zhang, J.~Zhang, B.~Zhou, E.~Ding, and J.~Wang, ``Gir: 3d gaussian inverse rendering for relightable scene factorization,'' 2023.

\bibitem{intrinsicImage}
H.~Barrow and J.~Tenenbaum, ``Recovering intrinsic scene characteristics from images,'' \emph{Recovering Intrinsic Scene Characteristics from Images}, 01 1978.

\bibitem{RaviInverseRendering}
\BIBentryALTinterwordspacing
R.~Ramamoorthi and P.~Hanrahan, ``A signal-processing framework for inverse rendering,'' in \emph{Proceedings of the 28th Annual Conference on Computer Graphics and Interactive Techniques}, ser. SIGGRAPH '01.\hskip 1em plus 0.5em minus 0.4em\relax New York, NY, USA: Association for Computing Machinery, 2001, p. 117–128. [Online]. Available: \url{https://doi.org/10.1145/383259.383271}
\BIBentrySTDinterwordspacing

\bibitem{SatoInverseRendering}
Y.~Sato, M.~Wheeler, and K.~Ikeuchi, ``Object shape and reflectance modeling from observation,'' \emph{proceedings of ACM Siggraph '97 (Computer Graphics)}, 04 1997.

\bibitem{InverseRenderingSurvey03}
G.~Patow and X.~Pueyo, ``{A Survey of Inverse Rendering Problems},'' \emph{Computer Graphics Forum}, vol.~22, no.~4, pp. 663--687, 2003.

\bibitem{kato2020differentiable}
H.~Kato, D.~Beker, M.~Morariu, T.~Ando, T.~Matsuoka, W.~Kehl, and A.~Gaidon, ``Differentiable rendering: A survey,'' \emph{arXiv preprint arXiv:2006.12057}, 2020.

\bibitem{azinovic2019inverse}
D.~Azinovic, T.-M. Li, A.~Kaplanyan, and M.~Nie{\ss}ner, ``Inverse path tracing for joint material and lighting estimation,'' in \emph{Proceedings of the IEEE/CVF CVPR}, 2019, pp. 2447--2456.

\bibitem{ndrmc}
J.~Hasselgren, N.~Hofmann, and J.~Munkberg, ``Shape, light, and material decomposition from images using monte carlo rendering and denoising,'' \emph{Advances in Neural Information Processing Systems}, vol.~35, pp. 22\,856--22\,869, 2022.

\bibitem{nicolet2021large}
B.~Nicolet, A.~Jacobson, and W.~Jakob, ``Large steps in inverse rendering of geometry,'' \emph{ACM Transactions on Graphics (TOG)}, vol.~40, no.~6, pp. 1--13, 2021.

\bibitem{luan2021unified}
F.~Luan, S.~Zhao, K.~Bala, and Z.~Dong, ``Unified shape and svbrdf recovery using differentiable monte carlo rendering,'' in \emph{Computer Graphics Forum}, vol.~40, no.~4.\hskip 1em plus 0.5em minus 0.4em\relax Wiley Online Library, 2021, pp. 101--113.

\bibitem{queau2018variational}
Y.~Qu{\'e}au, J.~M{\'e}lou, F.~Castan, D.~Cremers, and J.-D. Durou, ``A variational approach to shape-from-shading under natural illumination,'' in \emph{Energy Minimization Methods in Computer Vision and Pattern Recognition: 11th International Conference, EMMCVPR 2017, Venice, Italy, October 30--November 1, 2017, Revised Selected Papers 11}.\hskip 1em plus 0.5em minus 0.4em\relax Springer, 2018, pp. 342--357.

\bibitem{DVR}
M.~Niemeyer, L.~Mescheder, M.~Oechsle, and A.~Geiger, ``Differentiable volumetric rendering: Learning implicit 3d representations without 3d supervision,'' in \emph{Proceedings of the IEEE/CVF conference on computer vision and pattern recognition}, 2020, pp. 3504--3515.

\bibitem{IDR}
L.~Yariv, Y.~Kasten, D.~Moran, M.~Galun, M.~Atzmon, B.~Ronen, and Y.~Lipman, ``Multiview neural surface reconstruction by disentangling geometry and appearance,'' \emph{Advances in Neural Information Processing Systems}, vol.~33, pp. 2492--2502, 2020.

\bibitem{nvidiffrec}
J.~Munkberg, J.~Hasselgren, T.~Shen, J.~Gao, W.~Chen, A.~Evans, T.~M{\"u}ller, and S.~Fidler, ``Extracting triangular 3d models, materials, and lighting from images,'' in \emph{Proceedings of the IEEE/CVF CVPR}, 2022, pp. 8280--8290.

\bibitem{pbir}
C.~Sun, G.~Cai, Z.~Li, K.~Yan, C.~Zhang, C.~Marshall, J.-B. Huang, S.~Zhao, and Z.~Dong, ``Neural-pbir reconstruction of shape, material, and illumination,'' in \emph{Proceedings of the IEEE/CVF International Conference on Computer Vision}, 2023, pp. 18\,046--18\,056.

\bibitem{nerv}
P.~P. Srinivasan, B.~Deng, X.~Zhang, M.~Tancik, B.~Mildenhall, and J.~T. Barron, ``Nerv: Neural reflectance and visibility fields for relighting and view synthesis,'' in \emph{Proceedings of the IEEE/CVF CVPR}, 2021, pp. 7495--7504.

\bibitem{physg}
K.~Zhang, F.~Luan, Q.~Wang, K.~Bala, and N.~Snavely, ``Physg: Inverse rendering with spherical gaussians for physics-based material editing and relighting,'' in \emph{Proceedings of the IEEE/CVF CVPR}, 2021, pp. 5453--5462.

\bibitem{zhang2022modeling}
Y.~Zhang, J.~Sun, X.~He, H.~Fu, R.~Jia, and X.~Zhou, ``Modeling indirect illumination for inverse rendering,'' in \emph{Proceedings of the IEEE/CVF CVPR}, 2022, pp. 18\,643--18\,652.

\bibitem{mai2023neural}
A.~Mai, D.~Verbin, F.~Kuester, and S.~Fridovich-Keil, ``Neural microfacet fields for inverse rendering,'' in \emph{Proceedings of the IEEE/CVF International Conference on Computer Vision}, 2023, pp. 408--418.

\bibitem{neus}
P.~Wang, L.~Liu, Y.~Liu, C.~Theobalt, T.~Komura, and W.~Wang, ``Neus: Learning neural implicit surfaces by volume rendering for multi-view reconstruction,'' \emph{arXiv preprint arXiv:2106.10689}, 2021.

\bibitem{unisurf}
M.~Oechsle, S.~Peng, and A.~Geiger, ``Unisurf: Unifying neural implicit surfaces and radiance fields for multi-view reconstruction,'' in \emph{Proceedings of the IEEE/CVF International Conference on Computer Vision}, 2021, pp. 5589--5599.

\bibitem{gao2023relightable}
J.~Gao, C.~Gu, Y.~Lin, H.~Zhu, X.~Cao, L.~Zhang, and Y.~Yao, ``Relightable 3d gaussian: Real-time point cloud relighting with brdf decomposition and ray tracing,'' 2023.

\bibitem{wu2024deferredgs}
T.~Wu, J.-M. Sun, Y.-K. Lai, Y.~Ma, L.~Kobbelt, and L.~Gao, ``Deferredgs: Decoupled and editable gaussian splatting with deferred shading,'' 2024.

\bibitem{our2024sig3dgsdf}
K.~Ye, Q.~Hou, and K.~Zhou, ``3d gaussian splatting with deferred reflection,'' in \emph{SIGGRAPH Asia 2024 Conference Papers}, 2024, p. TBD.

\bibitem{cook1982reflectance}
R.~L. Cook and K.~E. Torrance, ``A reflectance model for computer graphics,'' \emph{ACM Transactions on Graphics (ToG)}, vol.~1, no.~1, pp. 7--24, 1982.

\bibitem{ibr}
S.~Chan, H.-Y. Shum, and K.-T. Ng, ``Image-based rendering and synthesis,'' \emph{IEEE Signal Processing Magazine}, vol.~24, no.~6, pp. 22--33, 2007.

\bibitem{whitted1980improved}
\BIBentryALTinterwordspacing
T.~Whitted, ``An improved illumination model for shaded display,'' \emph{Commun. ACM}, vol.~23, no.~6, p. 343–349, jun 1980. [Online]. Available: \url{https://doi.org/10.1145/358876.358882}
\BIBentrySTDinterwordspacing

\bibitem{nimier2022unbiased}
M.~Nimier-David, T.~M{\"u}ller, A.~Keller, and W.~Jakob, ``Unbiased inverse volume rendering with differential trackers,'' \emph{ACM Transactions on Graphics (TOG)}, vol.~41, no.~4, pp. 1--20, 2022.

\bibitem{DeferredRendering}
M.~Deering, S.~Winner, B.~Schediwy, C.~Duffy, and N.~Hunt, ``The triangle processor and normal vector shader: a vlsi system for high performance graphics,'' in \emph{Proceedings of SIGGRAPH '88}.\hskip 1em plus 0.5em minus 0.4em\relax New York, NY, USA: ACM, 1988, p. 21–30.

\bibitem{PRT}
P.-P. Sloan, J.~Kautz, and J.~Snyder, ``Precomputed radiance transfer for real-time rendering in dynamic, low-frequency lighting environments,'' in \emph{Proceedings of the 29th annual conference on Computer graphics and interactive techniques}, 2002, pp. 527--536.

\bibitem{sloan2008stupid}
P.-P. Sloan, ``Stupid spherical harmonics (sh) tricks,'' in \emph{Game developers conference}, vol.~9, 2008, p.~42.

\bibitem{hill2020physically}
S.~Hill, S.~McAuley, L.~Belcour, W.~Earl, N.~Harrysson, S.~Hillaire, N.~Hoffman, L.~Kerley, J.~Patry, R.~Piek{\'e} \emph{et~al.}, ``Physically based shading in theory and practice,'' in \emph{ACM SIGGRAPH 2020 Courses}, 2020, pp. 1--12.

\bibitem{stanfordorb}
Z.~Kuang, Y.~Zhang, H.-X. Yu, S.~Agarwala, S.~Wu, and J.~Wu, ``Stanford-orb: A real-world 3d object inverse rendering benchmark,'' 2023.

\bibitem{jin2023tensoir}
H.~Jin, I.~Liu, P.~Xu, X.~Zhang, S.~Han, S.~Bi, X.~Zhou, Z.~Xu, and H.~Su, ``Tensoir: Tensorial inverse rendering,'' in \emph{Proceedings of the IEEE/CVF Conference on Computer Vision and Pattern Recognition}, 2023, pp. 165--174.

\bibitem{arrighetti2017academy}
W.~Arrighetti, ``The academy color encoding system (aces): A professional color-management framework for production, post-production and archival of still and motion pictures,'' \emph{Journal of Imaging}, vol.~3, no.~4, p.~40, 2017.

\bibitem{envidr}
R.~Liang, H.~Chen, C.~Li, F.~Chen, S.~Panneer, and N.~Vijaykumar, ``Envidr: Implicit differentiable renderer with neural environment lighting,'' in \emph{Proceedings of the IEEE/CVF ICCV}, 2023, pp. 79--89.

\bibitem{walter2007microfacet}
B.~Walter, S.~R. Marschner, H.~Li, and K.~E. Torrance, ``Microfacet models for refraction through rough surfaces,'' in \emph{Proceedings of the 18th Eurographics conference on Rendering Techniques}, 2007, pp. 195--206.

\bibitem{zhou2005precomputed}
K.~Zhou, Y.~Hu, S.~Lin, B.~Guo, and H.-Y. Shum, ``Precomputed shadow fields for dynamic scenes,'' in \emph{ACM SIGGRAPH 2005 Papers}, 2005, pp. 1196--1201.

\bibitem{ramamoorthi2001relationship}
R.~Ramamoorthi and P.~Hanrahan, ``On the relationship between radiance and irradiance: determining the illumination from images of a convex lambertian object,'' \emph{JOSA A}, vol.~18, no.~10, pp. 2448--2459, 2001.

\bibitem{mip_nerf_360}
J.~T. Barron, B.~Mildenhall, D.~Verbin, P.~P. Srinivasan, and P.~Hedman, ``Mip-nerf 360: Unbounded anti-aliased neural radiance fields,'' in \emph{Proceedings of the IEEE/CVF CVPR}, 2022, pp. 5470--5479.

\end{thebibliography}
